\documentclass[11pt]{article}

\usepackage[final]{acl}

\usepackage{times}
\usepackage{latexsym}

\usepackage[T1]{fontenc}
\usepackage[utf8]{inputenc}

\usepackage{microtype}

\usepackage{inconsolata}

\usepackage{graphicx}
\usepackage{hyperref}
\usepackage{url}
\usepackage{graphicx}
\usepackage{adjustbox}
\usepackage{booktabs} % for professional tables
\usepackage{multirow}
\usepackage{comment}
\usepackage{amsmath,amsfonts, bm}

\newcommand\blfootnote[1]{% 
\begingroup 
\renewcommand\thefootnote{}\footnote{#1}% 
\addtocounter{footnote}{-1}% 
\endgroup 
}

\title{InComeS: Integrating Compression and Selection Mechanisms into LLMs for Efficient Model Editing\thanks{This work is partially supported by CityU Startup Grant (Grant No. 9382014) and the Singapore Ministry of Education (MOE) Academic Research Fund (AcRF) Tier 1 grant (Proposal ID: 24-SIS-SMU-002). Zhisong Zhang is supported by CityU Startup Grant (Grant No. 9382014). Yang Deng is support by the Lee Kong Chian Fellowship awarded by Singapore Management University.}}

\author{Shuaiyi Li$^1$\thanks{Work done during internship at Tencent AI Lab.}, Zhisong Zhang$^{2,\dagger}$, Yang Deng$^4$, \\ 
\textbf{Chenlong Deng$^3$, Tianqing Fang$^3$, Hongming Zhang$^3$}, \\
\textbf{Haitao Mi$^3$, Dong Yu$^3$, Wai Lam$^{1,\dagger}$} \\
  $^1$The Chinese University of Hong Kong, \\ 
  $^2$Department of Computer Science, City University of Hong Kong, \\ 
  $^3$Tencent AI Lab, $^4$Singapore Management University \\
  \texttt{\{sli, wlam\}@se.cuhk.edu.hk}, \texttt{zhisong.zhang@cityu.edu.hk}}

\begin{document}
\maketitle
\begin{abstract}
Although existing model editing methods perform well in recalling exact edit facts, they often struggle in complex scenarios that require deeper semantic understanding rather than mere knowledge regurgitation. Leveraging the strong contextual reasoning abilities of large language models (LLMs), in-context learning (ICL) becomes a promising editing method by comprehending edit information through context encoding. However, this method is constrained by the limited context window of LLMs, leading to degraded performance and efficiency as the number of edits increases. To overcome this limitation, we propose InComeS, a flexible framework that enhances LLMs’ ability to process editing contexts through explicit compression and selection mechanisms. Specifically, InComeS compresses each editing context into the key-value (KV) cache of a special token, enabling efficient handling of multiple edits without being restricted by the model’s context window. Furthermore, specialized cross-attention modules are added to dynamically select the most relevant information from the pool of special tokens, enabling adaptive and effective utilization of edit information. We conduct experiments on diverse model editing benchmarks with various editing formats, and the results demonstrate the effectiveness and efficiency of our method.
\blfootnote{$^\dagger$ Corresponding author.}
\end{abstract}

\section{Introduction}
\label{introduction}
Model editing, also known as knowledge editing, has seen rapid progress in recent years \cite{alphaedit, coachhook, wise, knowedit}. Its primary goal is to precisely integrate updated knowledge into a model, enabling targeted behavioral modifications while maintaining performance on unrelated tasks. Existing techniques have demonstrated strong performance in accurately recalling edited facts \cite{Yao, knowedit, edit-overfit}. However, they often struggle in more complex editing scenarios, such as multi-hop editing composition \cite{mquake, edit-overfit}, natural language editing \cite{dune}, and editing tasks that require reasoning and generalization \cite{ripple-effects, knowedit}. Moreover, recent studies \cite{edit-overfit} show that previous editing methods are prone to overfitting: they may assign excessively high probabilities to edited targets, which can distort the model’s responses to more complex or nuanced queries.

Leveraging the in-context learning (ICL) abilities of large language models (LLMs) provides a promising direction for addressing these problems. As LLMs continue to grow in size and capability, their ability to understand and utilize contextual information continues to improve. By incorporating all the editing information into the prefix contexts, ICL enables a simple, powerful, and flexible approach for employing updated knowledge in complex scenarios. However, this approach faces significant challenges as the number of edits increases. 
First, the finite context window restricts the maximum number of edits that can be included, and the computational cost of self-attention over long contexts leads to a sharp decline in \emph{efficiency}. 
Moreover, the effectiveness of ICL is constrained by the model’s ability to process extended contexts, and the retrieval \emph{accuracy} of the most relevant editing information also tends to decrease as the editing context grows.

To address these challenges, we introduce \textit{\textbf{InComeS}} 
\footnote{https://github.com/Syon-Li/InComeS}  
(\textbf{In}tegrating \textbf{Com}pr\textbf{e}ssion and \textbf{S}election Mechanisms), a novel framework for efficient and scalable model editing. InComeS adopts context compression techniques to condense the representation of each edit into the KV cache of special gist tokens, which can be cached and reused for computational efficiency. While gisting~\cite{gisting} was originally developed to compress single-input prompts, we extend this approach to handle multiple edits by further introducing a specialized selection mechanism. We further augment the model with cross-attention modules that allow each input token to attend to the compressed gist representations of edits, enabling fine-grained and adaptive selection of the most relevant information. Since each edit is compressed in parallel, our framework overcomes the limitations imposed by the context window, and the specialized selection modules can be learned to enhance retrieval accuracy.

We conduct experiments across a range of complex model editing settings, including multi-hop editing, natural language editing, and tasks requiring complicated reasoning. Experimental results demonstrate that InComeS outperforms existing editing methods, effectively handling diverse editing scenarios while offering efficiency gains.

\section{Preliminary}

Model editing~\cite{Yao, mend} aims to adjust a base model $\psi$ to a post-edited model $\psi'$ according to a set of editing information $\mathcal{T} = \{t_1,\dots,t_n\}$: $\psi' = \text{Edit}(\psi, \{t_1,\dots,t_n\})$.
% \begin{gather}
% \psi' = \text{Edit}(\psi, \{t_1,\dots,t_n\})
% \end{gather}
Here, ``$\text{Edit}$'' indicates the model editing method, while $\{t_1,\dots,t_n\}$ represents the knowledge pieces to be integrated. A typical example of editing information is query-label pair $t=(x, y)$, where the goal is for the edited model to produce $y$ in response to input $x$, even if the original model does not: $\psi(x) \neq y, \psi'(x) = y$.
% \begin{gather}
% \psi(x) \neq y, \quad \psi'(x) = y
% \end{gather}
When the editing set contains only a single piece of information ($|\mathcal{T}|=1$), this is known as single-instance editing. In contrast, batch editing refers to the scenario where multiple pieces of knowledge are updated simultaneously ($|\mathcal{T}|>1$). Batch editing is particularly practical in real-world applications, where simultaneously updating several edits is often required. In these scenarios, it will be more efficient to integrate them into the model in a single operation.

In practice, editing information can take various forms beyond simple query-label pairs. For instance, multiple related edits can be combined to enable multihop editing, or updated knowledge may be provided as a paragraph of natural language text. In such scenarios, many traditional editing methods may struggle to produce the desired outcomes, since they are not designed to handle these diverse types of editing information. In contrast, in-context learning (ICL) approaches, where editing information is simply concatenated as contextual prefixes, offer a straightforward yet powerful solution: $\text{Edit}_\text{ICL}(\psi, \{t_1,\dots,t_n\})(x) = \psi(t_1,\dots,t_n, x)$.
% \begin{gather}
% \text{Edit}_\text{ICL}(\psi, \{t_1,\dots,t_n\})(x) = \psi(t_1,\dots,t_n, x)
% \end{gather}
By leveraging the LLM’s ability to understand and reason over context, ICL can naturally accommodate a wide range of editing scenarios. Nevertheless, ICL is constrained by the context window of LLMs, and its accuracy and efficiency tend to decline when processing larger batches of edits.

\section{Method}

\begin{figure*}[t]
\centering
\includegraphics[width=0.9\textwidth]{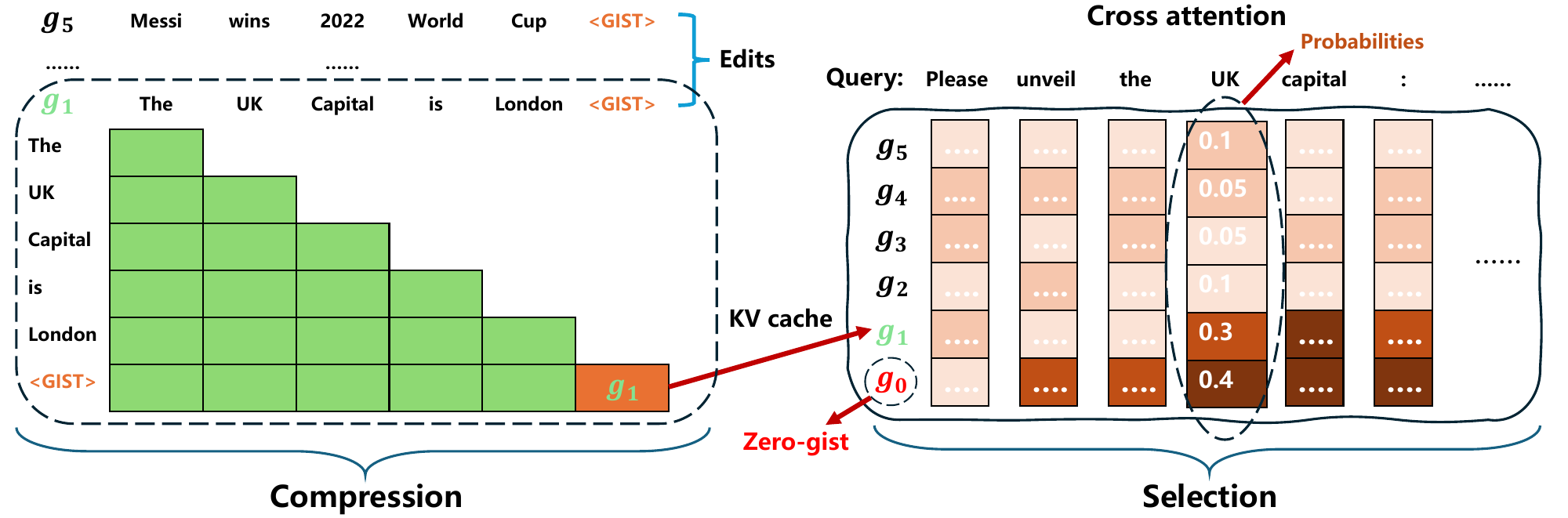}
\caption{An overview of \textit{InComeS}. At the compression stage, each edit is individually condensed into KV cache representations of a gist token. These representations are integrated into the model via selection through the cross-attention modules. A special zero-gist token is included alongside the cached gists from actual edits, allowing the model to have the option to ``select nothing.'' Note that the compression and integration steps are performed separately, but both use the same underlying model.}
\label{fig.method}
% \vspace{-0.4cm}
\end{figure*}

% ----- ZSTART ----- %
In this work, we aim to enhance the ICL-based editing approach to better understand multiple edits and accurately extract relevant information from the edit batch. Given a batch of editing information set $\mathcal{T} = \{t_1,\dots,t_n\}$, an input query $x$, and the subset of its related\footnote{We define related edits as those editing pieces that the model should reference when answering the query.} edits $\{t_i | i \in \mathfrak{R}(x)\}$, we hope that our model can answer the query as effectively as a vanilla LM provided only with the relevant edits (ignoring the irrelevant editing information): $\psi' = \text{Edit}(\psi, \{t_1,\dots,t_n\})  \approx \text{Edit}(\psi, \{t_i | i \in \mathfrak{R}(x)\})$.
% \begin{gather}
% \label{eq.goal}
% \psi' = \text{Edit}(\psi, \{t_1,\dots,t_n\})  \approx \text{Edit}(\psi, \{t_i | i \in \mathfrak{R}(x)\})
% \end{gather}
% where $\psi$ is a vanilla LM before editing, $\text{Edit}$ is the editing operation, and $\psi'$ is the LM edited with the editing contexts.

To enable accurate and efficient batch editing, we propose \textit{\textbf{InComeS}} %(Fig. \ref{fig.method}), 
an ICL-based approach that integrates both compression and selection mechanisms into the LMs. First, we adopt gist-based edit compression, condensing each editing information into the KV cache representations of one special (gist) token. Furthermore, we introduce parallel-context cross-attention modules that allow ordinary tokens to attend to these compressed gist representations. These modules serve as soft selectors to dynamically identify the most relevant information for the current input. This strategy can effectively mitigate the limitations imposed by context window sizes and enhance the model's ability to precisely capture editing information.
% \vspace{-0.2cm}
% ----- ZEND ----- %

% ----- ZSTART ----- %

\subsection{Edit Compression}
We adopt the concept of gisting \cite{gisting}, which is originally developed to compress input prompts into the representations of an extra, specially inserted token (the gist token). The condensed gist activation serves the same function as the original prompt and can be cached for later reuse, thereby improving computational and memory efficiency. While the original work primarily focuses on instruction tuning, we extend this idea to edit compression.

For each editing information piece $t_i$, represented as a sequence of tokens $t_i^0, t_i^1, \dots, t_i^n$, we append a special gist token $t_g$ to the end of the sequence and feed it into the LM. After encoding, we discard the original edit tokens and retain only the gist's representations (KV caches) for each edit. Notably, each edit context is encoded independently, allowing us to efficiently handle an arbitrary number of edits. This approach is highly flexible and accommodates edits of varying lengths and formats.

After edit compression, the edit information $t_1, t_2, \dots, t_n$ is converted into their corresponding gist KV representations\footnote{For brevity, we present the representations and operations for a single layer.} $(gK_1, gV_1), (gK_2, gV_2), \dots, (gK_n, gV_n)$. Importantly, we use the same LM targeted for editing to encode and compress the edit information, ensuring that the subsequent information selection process is seamless and well-aligned with the model’s internal representations.
% \vspace{-0.4cm}

\subsection{Edit Selection}
After compressing the edit contexts, we obtain a pool of gist representations for the batch of edits. To integrate this information into the model, we introduce additional cross-attention modules that enable input tokens to attend to the edit representations. Since these representations are stored as KV caches, we leverage a similar attention mechanism to incorporate the edit information. Formally, given a token’s query state $q$, the cross-attention is computed as: $o_{cross} = attention(q, \{gK_0, gK_1, gK_2, \dots, gK_n\}, \allowbreak \{gV_0, gV_1, gV_2, \dots, gV_n\})$.
% \begin{equation}
% \begin{split}
% o_{cross} = &attention(q, \\
% &\{gK_0, gK_1, gK_2, \dots, gK_n\}, \\
% &\{gV_0, gV_1, gV_2, \dots, gV_n\})
% \end{split}
% \end{equation}
We finally add the cross-attention outputs to the self-attention outputs for information aggregation.

Since tokens are not required to always attend to the edit information, we further introduce a zero-gist ($g_0$ in Figure~\ref{fig.method}) to allow the model to attend to ``nothing'' when appropriate. For the zero-gist, we use learnable parameters for the key vectors $gK_0$ and assign fixed zero vectors to the value $gV_0$. This design allows the model to flexibly select relevant information as needed during sequence prediction.

\subsection{Meta Training}

\begin{figure*}[t]
\centering
\includegraphics[width=0.9\textwidth]{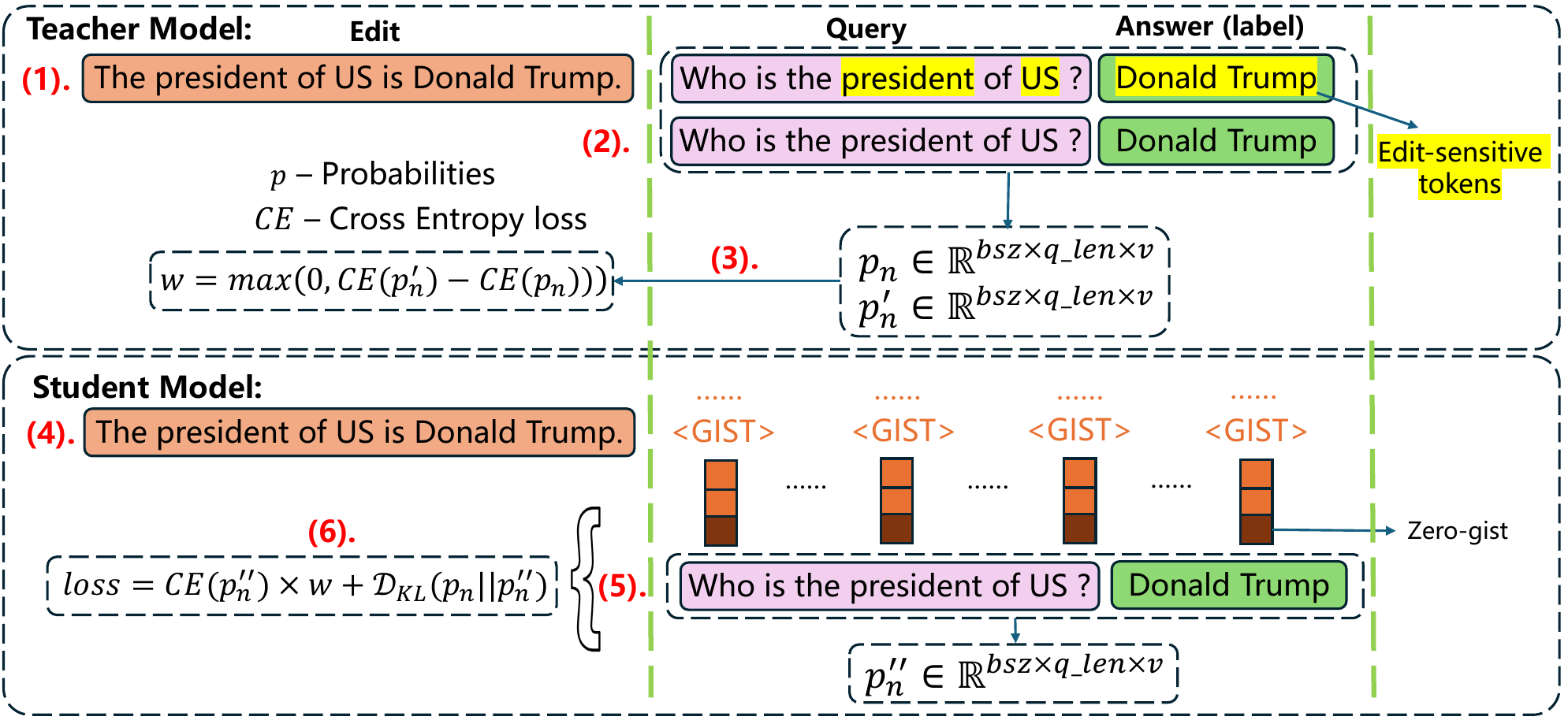}
\caption{An overview of the one-time meta training of \textit{InComeS}. The teacher model performs two forward passes: one with edit-contextualized input (1) and one with uncontextualized input (2). The cross-entropy between the outputs of (1) and (2) is used to compute a customized weight (3). The student model then compresses the edit information into KV representations using gist tokens (4). These KV caches are used to supply edit-relevant information to the query tokens (5). The final loss is computed as the sum of weighted cross entropy and KL divergence (6).}
\label{fig.training}
% \vspace{-0.4cm}
\end{figure*}

% ----- ZSTART ----- %

Since vanilla LMs lack explicit mechanisms for context compression and selection, we perform continued training (Figure~\ref{fig.training}) to enhance pre-trained LMs with these capabilities. Our main goal is to ensure that the compressed gist representations serve as effective substitutes for the original editing information. To achieve this, it is essential to distinguish between edit-sensitive tokens, whose losses change significantly when editing context is given, and edit-insensitive tokens, which can be predicted accurately from local context alone and do not depend on edit information. This distinction is captured by employing a customized token weighting scheme:
\begin{gather}
\begin{split}
w_{x_i} =& \max(0, \ CE(x_i|x_0, \dots, x_{i-1}) - \\
&CE(x_i|\{t_i | i \in \mathfrak{R}(x)\}, x_0, \dots, x_{i-1}))
\label{eq.token weights}
\end{split}
\end{gather}
where the $CE$ is the cross entropy and $\mathfrak{R}(x)$ represents the subset of related edits.
Here, the token weight is the difference between the edit-conditioned and edit-unconditioned losses. This scheme increases the weights of edit-sensitive tokens to encourage the model to learn to retrieve information from the compressed edits. The loss differences are calculated with a teacher model, which is the original, unedited version of the target LM.

In addition to token reweighting, we also adopt knowledge distillation \cite{distillation} to transfer the teacher model's knowledge about the edit information into the target model. Specifically, we apply the KL divergence to align the output distributions of the gist-contextualized student model with those of the edit-contextualized teacher model:
\begin{gather}
\begin{split}
KL_{x_i} =& D_{KL}(p_T(x_i|\{t_i | i \in \mathfrak{R}(x)\}, x_0, \dots, x_{i-1}) \ || \ \\
& p_S(x_i|\{g_1, \dots, g_n\}, x_0, \dots, x_{i-1}))
\label{eq.kl loss}
\end{split} \\
loss_{x_i} = w_{x_i} \cdot CE(x_i) + KL_{x_i}
\end{gather}
Here, $g_1, \dots, g_n$ denote the cached gists for all the edits. We apply the token reweighting only to the vanilla cross-entropy term in our final loss, since we found that it would degrade effective learning of "attend-to-nothing" behavior if combined with the KL part. The explicit training details are provided in Appendix \ref{appendix.training}.

\begin{table}[t]
\footnotesize
    \centering
    \setlength{\tabcolsep}{1mm}{
    \begin{adjustbox}{max width=\linewidth}
    \begin{tabular}{lcc}
    \toprule
    Method & Single Edit & Batch Edit \\ 
     \midrule
     \multicolumn{3}{c}{Llama-3.2-1B} \\
     \midrule
     Base & 38.96 & 38.96\\
     FT-M & 54.71 & 48.71 \\
     LoRA \cite{LoRA} & 65.56 & 48.44 \\
     ROME \cite{rome} & 3.43 & - \\
     R-ROME \cite{r-rome} & 5.18 & - \\
     MEMIT \cite{memit} & 34.03 & 24.76 \\
     EMMET \cite{emmet} & 5.58 & 14.85 \\
     GRACE \cite{grace} & 6.94 & 2.26 \\
     SERAC \cite{serac} & 38.97 & 39.01 \\
     MEND \cite{mend} & 36.41 & 31.91 \\
     RECIPE \cite{recipe} & 55.49 & 45.81 \\
     MeLLo \cite{mquake} & 53.09 & 44.10 \\
     MELO \cite{melo} & 40.33 & 41.47 \\
     ATBIAS \cite{atbias}  & 60.31 & 50.68 \\
     DR-IKE \cite{dr-ike} & 49.44 & 38.67 \\
     AlphaEdit \cite{alphaedit} & 51.93 & 41.98 \\
     ICE \cite{ice} & 52.81 & 45.38 \\
     ICL & 56.62 & 47.43 \\
     \textbf{InComeS} & \textbf{71.99} & \textbf{53.15} \\
     \midrule
     \multicolumn{3}{c}{Qwen2.5-7B} \\
     \midrule
     Base & 39.61 & 39.61 \\
     FT-M & 73.24 & 49.91 \\
     LoRA \cite{LoRA} & 33.42 & 21.25 \\
     ROME \cite{rome} & 8.64 & - \\
     R-ROME \cite{r-rome} & 7.01 & - \\
     MEMIT \cite{memit} & 40.78 & 40.20 \\
     EMMET \cite{emmet} & 30.28 & 39.91 \\
     GRACE \cite{grace} & 14.15 & 5.85 \\
     SERAC \cite{serac} & 56.15 & 40.69 \\
     MEND \cite{mend} & 36.61 & 36.51 \\
     RECIPE \cite{recipe} & 58.55 & 46.58 \\
     MeLLo \cite{mquake} & 55.50 & 47.34 \\
     MELO \cite{melo} & 40.49 & 42.20 \\
     ATBIAS \cite{atbias} & 51.17 & 45.00 \\
     DR-IKE \cite{dr-ike} & 55.61 & 41.37 \\
     AlphaEdit \cite{alphaedit} & 53.71 & 50.36 \\
     ICE \cite{ice} & 67.30 & 49.17 \\
     ICL & \textbf{73.74} & 49.61 \\
     \textbf{InComeS} & 71.41 & \textbf{52.17} \\
     \midrule
    \bottomrule
    \end{tabular}
    \end{adjustbox}}
    \caption{Results on MQuAKE \cite{mquake}. The full version can be found in Table \ref{tab:mquake-appendix}.}
    \label{tab:mquake}
    % \vspace{-0.4cm}
\end{table}

\begin{table}[t]
\footnotesize
    \centering
    \setlength{\tabcolsep}{1mm}{
    \begin{adjustbox}{max width=\textwidth}
    \begin{tabular}{lcc}
    \toprule
    Method & Single Edit & Batch Edit \\ 
     \midrule
     \multicolumn{3}{c}{Llama-3.2-1B} \\
     \midrule
     Base & 48.48 & 48.48 \\
     FT-M & 48.83 & 47.89 \\
     LoRA \cite{LoRA} & 47.08 & 49.08 \\
     SERAC \cite{serac} & 49.45 & 44.87 \\
     ICE \cite{ice} & 48.49 & 48.76 \\
     ICL & 57.05 & 50.69 \\
     \textbf{InComeS} & \textbf{57.59} & \textbf{53.45} \\
     \midrule
     \multicolumn{3}{c}{Qwen2.5-7B} \\
     \midrule
     Base & 55.47 & 55.47 \\
     FT-M & 60.44 & 56.78 \\
     LoRA \cite{LoRA} & 55.54 & 57.21 \\
     SERAC \cite{serac} & 53.80 & 48.99 \\
     ICE \cite{ice} & 55.30 & 46.66 \\
     ICL & 55.47 & 57.75 \\
     \textbf{InComeS} & \textbf{65.81} & \textbf{63.37} \\
     \midrule
    \bottomrule
    \end{tabular}
    \end{adjustbox}}
    \caption{Results on DUNE \cite{dune}. Detailed version can be found in Table \ref{tab:dune-appendix}.}
    \label{tab:dune}
    % \vspace{-0.4cm}
\end{table}

\begin{table}[t]
\footnotesize
    \centering
    \setlength{\tabcolsep}{1mm}{
    \begin{adjustbox}{max width=\linewidth}
    \begin{tabular}{lcccc}
    \toprule
    \multirow{2}{*}{Method} & \multicolumn{2}{c}{WikiData$_{cf}$} & \multicolumn{2}{c}{ZsRE-extended} \\ 
    \cmidrule(lr){2-3}\cmidrule(lr){4-5}
    
     & Single & Batch & Single & Batch \\
     \midrule
     \multicolumn{5}{c}{Llama-3.2-1B} \\
     \midrule
     Base & 19.73 & 19.73 & 40.17 & 40.17 \\
     FT-M & 53.43 & 47.51 & 62.80 & 54.84 \\
     LoRA \cite{LoRA} & 52.87 & 43.84 & 57.43 & 44.85 \\
     ROME \cite{rome} & 40.44 & - & 46.04 & - \\
     MEMIT \cite{memit} & 46.34 & 23.51 & 43.26 & 25.68 \\
     MEND \cite{mend}& 24.98 & 21.06 & 37.53 & 30.77 \\
     GRACE \cite{grace} & 14.33 & 10.51 & 12.73 & 10.91 \\
     IKE \cite{ike} & 45.55 & - & 57.39 & - \\
     SERAC \cite{serac} & 60.56 & 40.45 & 66.59 & 51.61 \\
     ICL & \textbf{65.81} & 44.75 & 62.19 & 51.58 \\
     \textbf{InComeS} & 65.15 & \textbf{45.66} & \textbf{70.70} & \textbf{52.23} \\
     \midrule
     \multicolumn{5}{c}{Qwen2.5-7B} \\
     \midrule
     Base & 21.46 & 21.46 & 43.86 & 43.86 \\
     FT-M & 49.39 & 43.13 & 50.04 & 46.41 \\
     LoRA \cite{LoRA} & 37.04 & 31.65 & 28.61 & 24.13 \\
     ROME \cite{rome} & 40.25 & - & 50.43 & - \\
     MEMIT \cite{memit} & 43.58 & 39.85 & 53.23 & 49.97 \\
     MEND \cite{mend} & 20.45 & 15.29 & 43.26 & 38.83 \\
     GRACE \cite{grace} & 25.60 & 18.55 & 14.35 & 11.25 \\
     IKE \cite{ike} & 65.33 & - & 70.17 & - \\
     SERAC \cite{serac} & 51.12 & 41.26 & 62.41 & 52.63 \\
     ICL & \textbf{66.99} & \textbf{51.66} & 66.10 & 60.57 \\
     \textbf{InComeS} & 66.69 & 47.93 & \textbf{75.63} & \textbf{61.22} \\
     \midrule
    \bottomrule
    \end{tabular}
    \end{adjustbox}}
    \caption{Portability results on WikiData$_{counterfact}$ \cite{ripple-effects, knowedit} and ZsRE-extended \cite{knowedit, Yao}. Full results can be found at Table \ref{tab:portability-result-appendix}.}
    \label{tab:portability-result}
    % \vspace{-0.4cm}
\end{table}

\section{Experiments}

\subsection{Experiment setting}

\paragraph{Datasets \& Evaluation Metrics}
To verify the effectiveness of our method in complex editing scenarios, we conduct experiments on five popular datasets in model editing: the dataset for multi-hop editing MQuAKE \cite{mquake}, the natural language editing dataset DUNE \cite{dune}, the extended version of ZsRE \cite{Yao, knowedit}, which adds a portability test set to the original ZsRE \cite{zsre}, and the dataset containing ripple effect samples $\text{WikiData}_{counterfact}$ \cite{ripple-effects, knowedit}. We report edit success rate and portability for the extend-ZsRE and $\text{WikiData}_{counterfact}$, the results for 2, 3, and 4 edits for MQuAKE \cite{mquake} and new information, scientific reasoning, and debiasing for DUNE \cite{dune}.
% To facilitate the later analysis, we define the edit batch size as the number of edit queries, and the gist batch size as the number of required gist KV caches for a corresponding number of edit queries.
More details about the datasets and evaluation metrics can be found in the Appendix \ref{appendix.ds} and Appendix \ref{appendix.eval-metric}, respectively.
% \vspace{-0.4cm}

\paragraph{Baselines}
For baselines, we select representative methods demonstrated to be powerful in relevant surveys \cite{Yao, knowedit}. For methods that directly edit the model weights, we include ROME \cite{rome}, R-ROME \cite{r-rome}, MEMIT \cite{memit}, EMMET \cite{emmet}, AlphaEdit \cite{alphaedit}; for methods that adopt explicit external memory, we include SERAC \cite{serac}, IKE \cite{ike}, and DR-IKE \cite{dr-ike}; for methods that requires meta-training (like InComeS) or use implicit external memory (stores activations or nerons, etc), we adopt MEND \cite{mend}, GRACE \cite{grace}, KN \cite{kn}, MELO \cite{melo} and RECIPE \cite{recipe}. We also include some ICL-related methods like ICE \cite{ice} and ATBIAS \cite{atbias}, and some traditional but powerful methods, like fine-tuning,  LoRA \cite{LoRA}, and ICL, which directly concatenates all the edits as the prefix context. While some similarities exist between our method and RAG, they vary considerably in problem setting and methodology. A detailed analysis is given in Appendix \ref{appendix.RAG-InComeS}.
We choose two representative open-source models for evaluation: Llama-3.2-1B \footnote{https://huggingface.co/meta-llama/Llama-3.2-1B} and Qwen2.5-7B \cite{qwen2} (More results about Qwen3-8B-base is in Appendix \ref{appendix.more-results-on-qwen3}). Unless otherwise specified, we adopt an edit batch size of 100 for batch editing.
More details on the baseline implementation can be found in the Appendix \ref{appendix.baseline-details}.

\subsection{Main results}

\paragraph{Multi-hop edits}
We test our method on MQuAKE for the multiple-hop edit scenario, where the models are required to check multiple edits to answer each query. Because of this requirement, we mainly compare with methods designed to support batch or sequential editing. Table \ref{tab:mquake} presents our main results, which demonstrate the effectiveness of InComeS in both single-editing and batch-editing scenarios. In addition, InComeS surpasses ICL in all metrics except %single 2-edits and 3-edits for Qwen2.5-7B, 
the single editing setting for Qwen2.5-7B, which shows that our method can effectively select relevant information from the editing contexts. Interestingly, single-edit specialized methods (such as ROME) collapse even in the single multi-hop query setting, revealing their incapability to handle complex editing scenarios. To verify the effectiveness of our method on different model scales, we conduct further analysis on the performance gain of different model scales in Appendix \ref{appendix.model-scale-results}.

\paragraph{Natural language edits}
One of our method's advantages is its flexibility in handling a variety of editing contexts with different formats. Unlike many traditional editing methods like ROME \cite{rome} and MEMIT \cite{memit}, which require the input to follow the triplet-like fact statement format, InComeS can take edits in free-text forms without explicitly labeled subjects and objects. To verify our method's capability for such scenarios, we adopt the DUNE dataset, which includes \textit{natural-language form edits}, and the results are shown in Table \ref{tab:dune}. Following the original paper of DUNE \cite{dune}, we include fine-tuning, LoRA, SERAC \cite{serac}, and ICL as our baselines. The result confirms our method's capability to handle natural language edits. Interestingly, the raw model itself is a strong baseline in the batch editing scenario, which may demonstrate the fast-evolving model capabilities over the years.

\paragraph{Evaluation on portability}
We further evaluate our method on two popular editing datasets that require reasoning abilities: $Wiki_{counterfact}$ \cite{ripple-effects, knowedit} and the extended ZsRE \cite{Yao, knowedit}. Table~\ref{tab:portability-result} shows the results. Our primary focus is on portability, as it serves as the most representative metric for assessing a model’s comprehensive understanding of the editing information. Overall, our method achieves performance comparable to ICL and consistently outperforms other baselines. More analysis about the detailed results is given in Appendix \ref{appendix.edit-success-metric}.

\begin{figure}[t]
\centering
\includegraphics[width=0.5\textwidth]{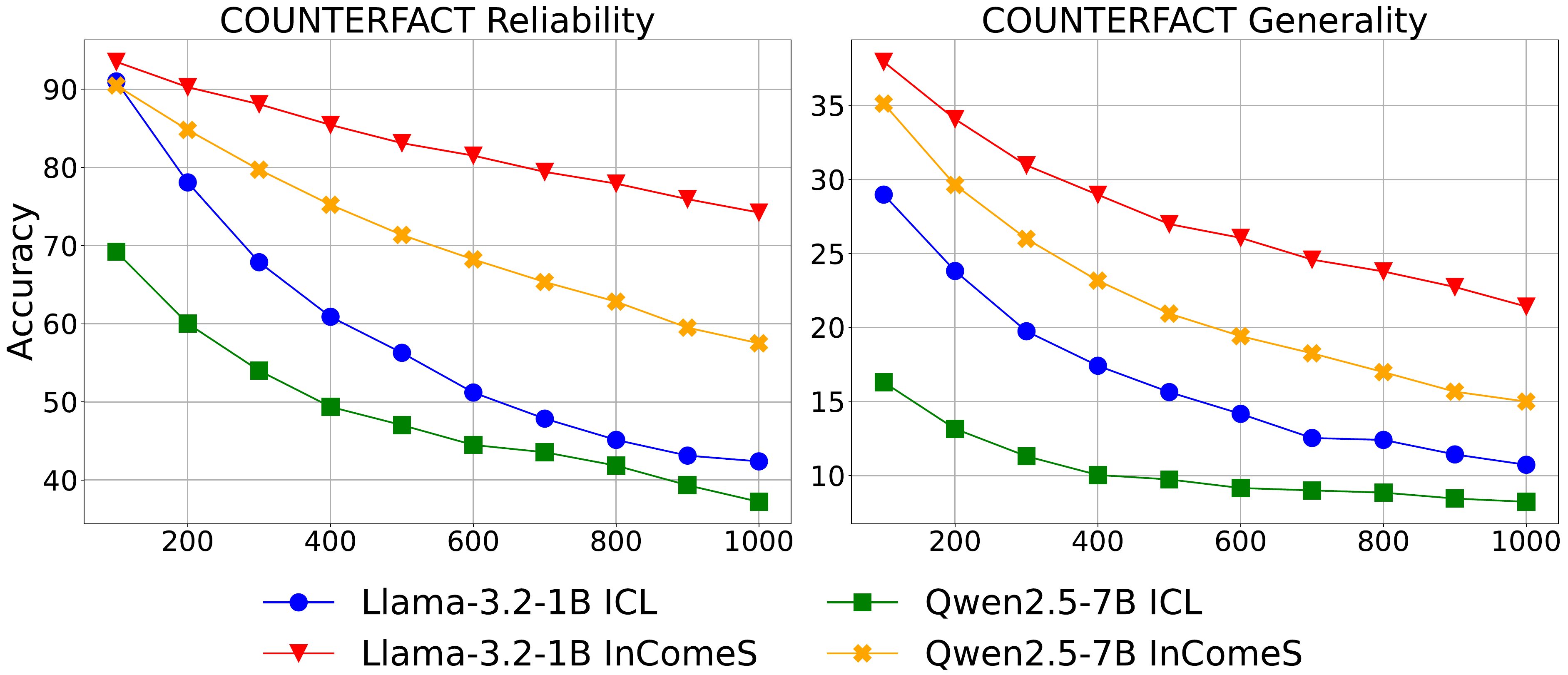}
\caption{Scaling-up analysis. We compare InComeS and ICL by varying the number of edits, as indicated on the $x$-axis.}
\label{fig.scale_up_context}
\end{figure}

\paragraph{Scaling up contexts}
\label{scale-up-contexts}
We further provide a scaling-up analysis to illustrate our method's ability to generalize to larger numbers of edits, which is the main motivation of our modification over the ICL baseline. For this analysis, we use the COUNTERFACT dataset \cite{memit}, as it provides a sufficient number of editing instances. We vary the number of edits from 100 to 1000, resulting in total token counts ranging from approximately $1.2K$ to $12K$. The results are shown in Figure~\ref{fig.scale_up_context}, which shows that InComeS consistently outperforms ICL, though the base models have already been pretrained over long contexts \cite{qwen2}. This finding suggests that the vanilla attention mechanism alone is insufficient to effectively comprehend and precisely select the required information from the context in complex editing scenarios. In contrast, our method demonstrates greater potential for handling large-scale edits through the unified compression and selection mechanism. Further scaling up experiments could be found in Appendix \ref{appendix.further-scaling-up-contexts}.

\begin{table}[t]
\footnotesize
    \centering
    \setlength{\tabcolsep}{1.5mm}{
    \begin{adjustbox}{max width=\textwidth}
    \begin{tabular}{lr}
    \toprule
     \multicolumn{2}{c}{Llama-3.2-1B} \\
     \midrule
     Method & Time (seconds) \\
     \midrule
     InComeS-Compression & \textbf{0.0326} \\
     ICL-Prefilling & 0.8934 \\
     FT-M & 3.4124 \\
     LoRA \cite{LoRA} & 33.5412 \\
     MEMIT \cite{memit} & 112.2238 \\
     EMMET \cite{emmet} & 158.6524 \\
     \midrule
     \multicolumn{2}{c}{Qwen2.5-7B} \\
     \midrule
     InComeS-Compression & \textbf{0.1071} \\
     ICL-Prefilling & 0.9082 \\
     FT-M & 28.6132 \\
     LoRA \cite{LoRA} & 122.1876 \\
     MEMIT \cite{memit} & 423.4823 \\
     EMMET \cite{emmet} & 512.4235 \\     
    \bottomrule
    \end{tabular}
    \end{adjustbox}}
    \caption{Measured time (seconds) for 100 edits.}
    \label{tab:time-analysis-editing}
    % \vspace{-0.4cm}
\end{table}

\begin{table}[t]
\footnotesize
    \centering
    \setlength{\tabcolsep}{1.5mm}{
    \begin{adjustbox}{max width=\textwidth}
    \begin{tabular}{lc}
    \toprule
     Method & Time (seconds) \\
     \midrule
     \multicolumn{2}{c}{Encoding w/ $1k$} \\
     \midrule
     InComeS-Compression & \textbf{0.2108} \\
     ICL-Prefilling & 1.0413 \\
     \midrule
     \multicolumn{2}{c}{Decoding w/ $1k$} \\
     \midrule
     InComeS-Selection & \textbf{0.0274} \\
     ICL-Generation (with prefilled cache) & 0.1555 \\
     \midrule
     \multicolumn{2}{c}{Encoding w/ $2k$} \\
     \midrule
     InComeS-Compression & \textbf{0.4051} \\
     ICL-Prefilling & 1.2165 \\
     \midrule
     \multicolumn{2}{c}{Decoding w/ $2k$} \\
     \midrule
     InComeS-Selection & \textbf{0.0297} \\
     ICL-Generation (with prefilled cache) & 0.3545 \\  
    \bottomrule
    \end{tabular}
    \end{adjustbox}}
    \caption{Scaled efficiency comparison (seconds) between InComeS and ICL using Llama-3.2-1B.}
    \label{tab:time-analysis-encoding-decoding}
\end{table}

\subsection{Efficiency Analysis}
% \vspace{-0.3cm}

Finally, we present the efficiency analysis for our method. By default, the individual edit length is around 10 to 11. We first compare the efficiency of our method with the efficiency of other knowledge editing methods. Table~\ref{tab:time-analysis-editing} reports the time required to perform 100 edits for each method. Our method has significantly better efficiency than the other presented editing methods.
%better than traditional editing methods and further improves over ICL by 17\% to 27\%. 
Additionally, compared to ICL, our approach only needs to maintain the KV cache of the gist representations from the deeper half layers, resulting in substantially lower memory cost.
To verify our method's superiority in efficiency on long context, we further conduct experiments on scaled context length (Table \ref{tab:time-analysis-encoding-decoding}). The result demonstrates the efficiency advantage of our method in both the encoding and decoding stages. More detailed analysis about the efficiency including time and memory can be found in Appendix \ref{appendix.further-efficiency} and Appendix \ref{appendix.memory-bound-quantification}.
% \vspace{-0.4cm}

\subsection{Ablation study \& Analysis}

\begin{figure*}[t]
\centering
\includegraphics[width=0.8\textwidth]{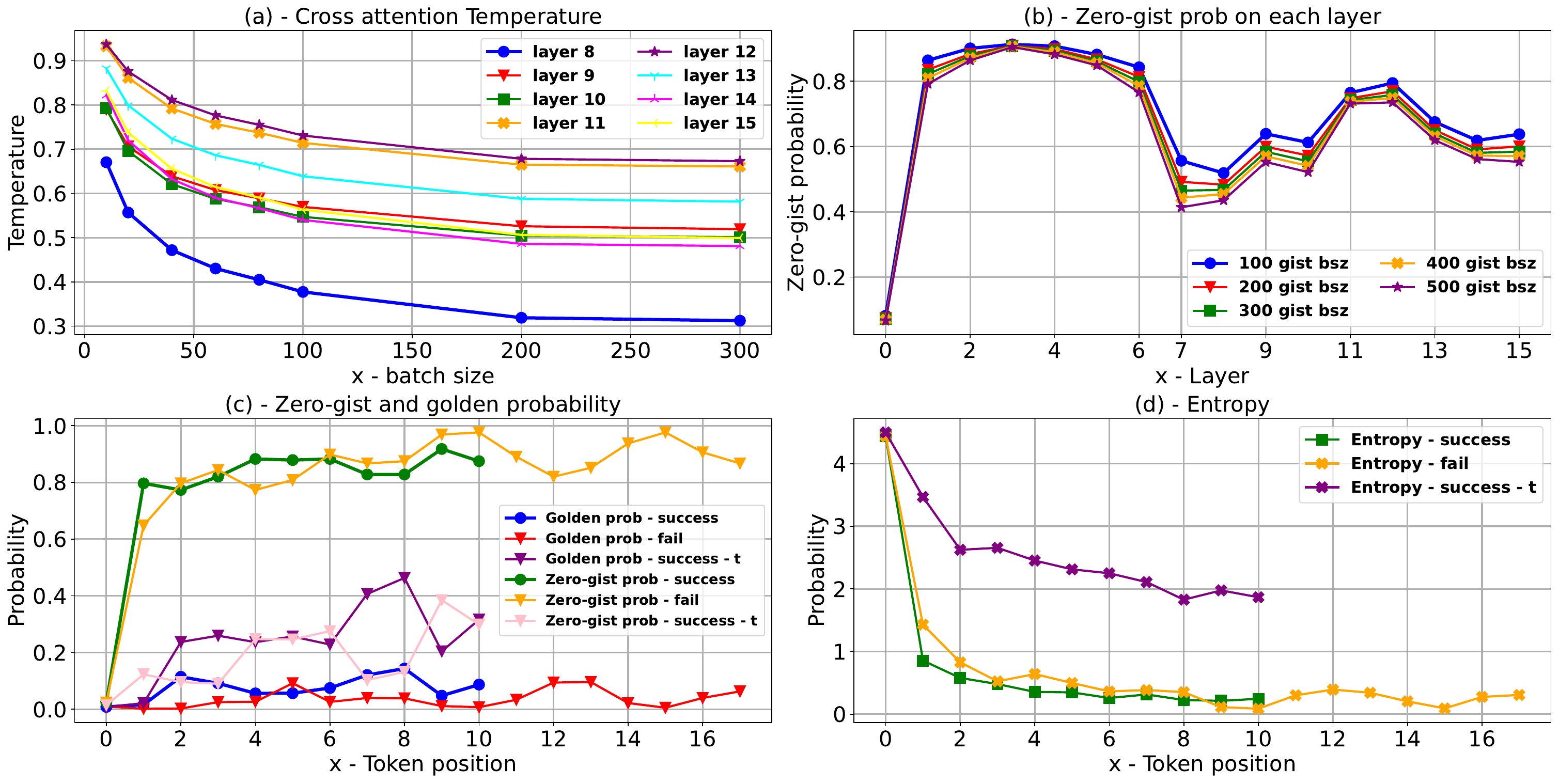}
% \vspace{-0.8cm}
\caption{Ablation and Analysis. (a) Experiments to investigate the desired temperature for cross-attention. (b) Investigation on the informativeness of the layers. (c) and (d) Study to reveal the selection pattern of the query tokens.}
\label{fig.Ablation-Analysis}
% \vspace{-0.4cm}
\end{figure*}

\begin{table}[t]
\footnotesize
    \centering
    \setlength{\tabcolsep}{1.5mm}{
    \begin{adjustbox}{max width=\linewidth}
    \begin{tabular}{lcc}
    \toprule
    \multirow{2}{*}{Method} & \multicolumn{2}{c}{MQuAKE} \\ 
    \cmidrule(lr){2-3}
    & Single editing & Batch editing \\ 
     \midrule
     \textbf{InComeS} & 71.99 & \textbf{53.15} \\
     - w/o zero-gist & 58.98 & 40.59 \\
     - w/ full model & \textbf{72.40} & 52.34 \\
     - w/o loss on query & 59.86 & 44.78 \\
     - w/ golden loss & 71.09 & 48.56 \\
     - w/o kl loss (Eq. \ref{eq.kl loss}) & 55.10 & 51.71 \\
     - w/o token weighting (Eq. \ref{eq.token weights}) & 55.02 & 52.63 \\
     \midrule
    \bottomrule
    \end{tabular}
    \end{adjustbox}}
    \caption{Ablation and Analysis experiments, the edit batch size is 100 for all results. Detailed results can be found in Table \ref{tab:ablation-analysis-appendix}.}
    \label{tab:ablation-analysis}
    % \vspace{-0.4cm}
\end{table}

\paragraph{Full model vs. Half model}
%Train a model with all layers used, and demonstrate that the first half of the layers mainly selects zero-gist.
We present the reason for our decision to use the KV cache from the second half of the model layers. To investigate this, we train a model using the KV cache from all layers and evaluate it on 1000 instances from ZsRE \cite{zsre}. We record the probabilities allocated to the zero-gist in the cross-attention modules, as shown in Fig. \ref{fig.Ablation-Analysis}b. The result shows that the zero-gist probabilities in layers 7-15 are generally lower than those in layers 1-6, and there is a notable drop in the zero-gist probability at layer 7. This suggests that, even when trained to use the full-model KV cache, the model mainly relies on information from deeper layers, since higher probabilities of the zero-gists indicate lower utilization of the actual edit contexts. A possible explanation is that more information is accumulated in the deeper layers, which aids both compression and selection processes. To verify our analysis, we also test the full layer trained model (see the ``w/ full model'' line in Table \ref{tab:ablation-analysis}). %The results shows a general decrease across all metrics except the single 4-edits case. 
The results show nearly no increase compared to the half model case (InComeS). Additionally, restricting the KV cache to only the second half of the model could provide efficiency benefits with lower memory and computation costs.

\paragraph{Deciding inference temperature}
\label{Temperature}
%1. Select a dataset, calculate the target entropy (only one edit). 2. Record the gist logits for 20, 40, 60, 80, 100, 120 edits and calculate the temperature. 3. Draw a figure for the edit batch size and the computed temperature.
Applying a small temperature to the gist cross-attention sharpens the probability distribution over the gist KV caches, which facilitates the model's ability to retrieve the correct information. We determine the appropriate temperature based on entropy, which has been shown to be an important factor in attention mechanisms \cite{atten-entropy}. Specifically, we aim to keep the cross-attention entropy close to its optimal value, which occurs when it only needs to attend to one edit. To achieve this, we select 1000 instances from ZsRE \cite{zsre} and calculate the entropy of the edit batch size 1. We then calculate the entropy for larger edit batch sizes (10, 20, 40, 60, 80, 100, 200, 300) and find the temperatures that align their entropy with the optimal case via gradient descent \footnote{Note that all entropy is calculated in a way that the query and answer part are just a copy of the context}. We report the calculated temperature in Figure~\ref{fig.Ablation-Analysis}a. As expected, the temperature decreases as the edit batch size increases, but interestingly, it gradually converges to a specific value. Specifically, layers 9, 10, and 14 finally converge to around 0.5. To encourage more decisive selection, we slightly lower this value and set the temperature to $T=0.45$.
% \vspace{-0.2cm}

\paragraph{Imposing golden loss on training}
As the golden gist representation is available for each training instance, it is natural to introduce an auxiliary loss to encourage correct selection in the cross-attention mechanism. We incorporate this additional loss in our experiments and report the result as ``w/ golden loss'' in Table~\ref{tab:ablation-analysis}. In the analysis in Figure~\ref{fig.Ablation-Analysis}, we use a suffix of ``- t'' to denote this setting. Incorporating the auxiliary loss leads the model to assign higher probabilities to the golden gist compared to training without this loss. Interestingly, it also increases the cross-attention entropy, probably because the model is explicitly encouraged to make selections during training. However, despite the increase in golden-gist probabilities, this approach does not yield clear performance improvements and even results in declines in some cases. This suggests that the model may develop its own context selection strategies, which do not always align with focusing all attention on the golden edit information.
% \vspace{-0.2cm}

\paragraph{Inclusion of zero gist}
The motivation for including the zero-gist mechanism is to ensure that context-independent tokens can bypass the influence of the edit contexts. To assess the impact of zero-gist, we train a model without this mechanism and evaluate it on MQuAKE \cite{mquake} (see the ``w/o zero-gist'' line in Table \ref{tab:ablation-analysis}). The results show a notable performance drop, suggesting that the cross-attention calculations may sometimes interfere with ordinary generation and our zero-gist strategy can mitigate this issue by allowing tokens to ``attend to nothing''.

\paragraph{Applying loss on queries}
\label{loss-beyond-labels}
By convention, instruction tuning only takes into account the loss for labels, excluding queries (Fig. \ref{fig.training}). In this section, we show that merely applying a loss on labels is not enough in our case. We train a model without the loss of queries and present its results in the Table~\ref{tab:ablation-analysis} (the line of ``w/o loss on query''). The absence of query loss results in a sharp decrease for multi-hop editing, suggesting that training on query tokens may improve the model's capability of combining information retrieval and reasoning.

\paragraph{Necessity of distillation}
We verify the importance of the distillation loss in this section. The result (see line "w/o kl loss" and "w/o token weighting" in Table \ref{tab:ablation-analysis}) shows a significant decrease when excluding any of the distillation components, indicating that the distillation plays an important role.

\paragraph{Information flow on tokens}
We further investigate the cross-attention patterns to understand how the model performs context selection. We measure the zero-gist and golden-gist probability (Figure~\ref{fig.Ablation-Analysis}c), and cross-attention entropy (Figure~\ref{fig.Ablation-Analysis}d) of each token from two representative examples containing a correctly ("success") and a wrongly ("fail") predicted instance using Llama-3.2-1B. As expected, the golden gist probability from the correctly predicted instance generally exceeds that of the failed one ("Golden prob - success" and "Golden prob - fail" line in Figure~\ref{fig.Ablation-Analysis}c). Notably, for all cases, the token at position zero allocates low probabilities to both golden and zero gists, while having high entropy, indicating that the model is ``taking the average'' of all gist representations at this beginning token. The dominance of zero-gist in later positions demonstrates that the model learns to "adaptively attend to edit information."

\paragraph{Analysis on cross-attention layers}
InComeS applies the cross attention only on the deeper half of the layers. Despite the relatively low zero-gist probability shown in Figure \ref{fig.Ablation-Analysis}b theoretically proves the necessity of the deeper half layers, we still conduct further analysis to verify their effectiveness (Table \ref{tab:analysis-on-deeper-half-layers}). The results show that each increase in the number of layers generally leads to a growth in the performance. This demonstrates the necessity of applying the whole deeper half layers.

\paragraph{More analysis} More analysis including side-effect analysis, etc. is provided in Appendix \ref{appendix.further-analysis}.

\section{Related Work}
The area of knowledge editing (or model editing) has experienced a thriving development in recent years. Researchers have explored various directions in this area.
One typical direction is to adopt external memory for the edits. The memory formats applied by different researchers are diverse. Methods like SERAC \cite{serac}, IKE \cite{ike}, DR-IKE \cite{dr-ike}, MeLLo \cite{mquake} adopt \textit{explicit non-parametric memory}, which stores specific edit instances, and a retriever that is responsible for recalling relevant edits from the memory. For example, IKE uses KNN, and SERAC applies a trained classifier. Another line of work, such as RECIPE \cite{recipe}, %CaliNET \cite{calinet}, T-Patcher \cite{t-patcher}, GRACE \cite{grace}, MELO \cite{melo}, KE \cite{ke}, MEND \cite{mend}, 
applies \textit{implicit parametric memory} to store the edits. CaliNET \cite{calinet}, T-Patcher \cite{t-patcher} embeds the knowledge into a fixed number of neurons and adds them to the model. GRACE \cite{grace} adopts a discrete key-value codebook with the value optimized for the desired knowledge. MELO \cite{melo} applies dynamic LoRA blocks and indexes them via an internal vector database. KE \cite{ke}, MEND \cite{mend} train a separate meta-model for editing. Another popular direction is to merge knowledge into the model directly. Methods like KN \cite{kn}, ROME \cite{rome}, R-ROME \cite{r-rome}, MEMIT \cite{memit}, PMET \cite{pmet}, CoachHooK \cite{coachhook}, and AlphaEdit \cite{alphaedit} perform editing by tweaking the located FFN part of the model directly. However, some studies reveal that these methods could potentially bring about side effects in the original model \cite{hurt-general-ability, Ocean-with-a-Spoon}, leaving the real effectiveness of these methods to be further investigated.

Unlike discrete prompts, soft prompts are learnable continuous vectors that are prepended to the input embedding sequence, enabling task-specific adaptation without modifying model weights. Prefix tuning \cite{prefix-Tuning} and prompt tuning \cite{prompt-tuning} insert trainable prefix activations or prompt embeddings into the input, achieving competitive performance with full fine-tuning while updating orders of magnitude fewer parameters. P-tuning \cite{p-tuning} further incorporates prompt encoders to model interdependencies among prompt tokens. These foundational works have inspired a wide range of subsequent innovations.

\section{Conclusion}
In this paper, we propose InComeS, a scalable model editing method that integrates compression and selection mechanisms directly into the LLMs. InComeS adopts a context compression technique to condense the editing context to KV representations on top of the introduced gist tokens and takes advantage of the compressed KVs to efficiently retrieve the relevant editing context information. Experiments on four different and complex editing settings demonstrate the superiority of our method for comprehensively editing. Further Analysis and ablations validate each component of InComeS and demonstrate the great efficiency and performance gain of our method.

\section*{Limitations}

\paragraph{Model scale \& Architecture}
Due to the limited computational resources, we only extend the model size to 7B and leave the larger model size to future work. We are aware that the original gisting work \cite{gisting} conducts experiments on three model architectures, i.e., encoder-decoder, encoder-only, and decoder-only. In this work, we determine to focus on the decoder-only autoregressive architecture as it is the structure used by most of the popular models nowadays \cite{qwen2, gpt-4, deepseek-v3}.  

\paragraph{Compression rate}
In this work, we maintain the compression rate to roughly around 12:1 (one edit, which contains around 12 tokens for all testing datasets except DUNE \cite{dune}, corresponds to one gist token), as one edit represents a fine-grained piece of information. However, we believe it is necessary to investigate the impact of lowering the compression rate, since it potentially helps extend the length of a single edit \cite{silver-bullet, unigist}.

\paragraph{Task variety}
InComeS can accept any input that follows natural language form. This flexibility gives it the potential to tackle many other tasks beyond knowledge editing. For example, long context language modeling, retrieval-augmented generation, etc. Due to the limited space of the main body of the paper, we first verify the effectiveness of our method on model editing and leave the investigation of other tasks to future work.

% \paragraph{Existence of scaling upper bound}
% Because all the gist has to be trained.

\bibliography{custom}

@inproceedings{Position-Really-Matters,
  author       = {Xianjun Yang and
                  Wei Cheng and
                  Xujiang Zhao and
                  Wenchao Yu and
                  Linda Ruth Petzold and
                  Haifeng Chen},
  editor       = {Luis Chiruzzo and
                  Alan Ritter and
                  Lu Wang},
  title        = {Position Really Matters: Towards a Holistic Approach for Prompt Tuning},
  booktitle    = {Findings of the Association for Computational Linguistics: {NAACL}
                  2025, Albuquerque, New Mexico, USA, April 29 - May 4, 2025},
  series       = {Findings of {ACL}},
  pages        = {8501--8523},
  publisher    = {Association for Computational Linguistics},
  year         = {2025},
  url          = {https://doi.org/10.18653/v1/2025.findings-naacl.474},
  doi          = {10.18653/V1/2025.FINDINGS-NAACL.474},
  bibsource    = {dblp computer science bibliography, https://dblp.org}
}

@inproceedings{MuppidiNB25,
  author       = {Ananth Muppidi and
                  Abhilash Nandy and
                  Sambaran Bandyopadhyay},
  editor       = {Wanxiang Che and
                  Joyce Nabende and
                  Ekaterina Shutova and
                  Mohammad Taher Pilehvar},
  title        = {Leveraging Self-Attention for Input-Dependent Soft Prompting in LLMs},
  booktitle    = {Proceedings of the 63rd Annual Meeting of the Association for Computational
                  Linguistics (Volume 2: Short Papers), {ACL} 2025, Vienna, Austria,
                  July 27 - August 1, 2025},
  pages        = {960--969},
  publisher    = {Association for Computational Linguistics},
  year         = {2025},
  url          = {https://doi.org/10.18653/v1/2025.acl-short.74},
  doi          = {10.18653/V1/2025.ACL-SHORT.74},
  bibsource    = {dblp computer science bibliography, https://dblp.org}
}

@article{ce-prompt,
  author       = {Wujian Yang and
                  Chunxu Jin and
                  Guanlin Chen and
                  Haotian Jin},
  title        = {CE-Prompt: enhance prompt expression stability by multiple understanding},
  journal      = {PeerJ Comput. Sci.},
  volume       = {11},
  pages        = {e3231},
  year         = {2025},
  url          = {https://doi.org/10.7717/peerj-cs.3231},
  doi          = {10.7717/PEERJ-CS.3231},
  bibsource    = {dblp computer science bibliography, https://dblp.org}
}

@inproceedings{adept,
  author       = {Pengwei Tang and
                  Xiaolin Hu and
                  Yong Liu},
  title        = {ADePT: Adaptive Decomposed Prompt Tuning for Parameter-Efficient Fine-tuning},
  booktitle    = {The Thirteenth International Conference on Learning Representations,
                  {ICLR} 2025, Singapore, April 24-28, 2025},
  publisher    = {OpenReview.net},
  year         = {2025},
  url          = {https://openreview.net/forum?id=fswihJIYbd},
  bibsource    = {dblp computer science bibliography, https://dblp.org}
}

@article{p-tuning,
  author       = {Xiao Liu and
                  Yanan Zheng and
                  Zhengxiao Du and
                  Ming Ding and
                  Yujie Qian and
                  Zhilin Yang and
                  Jie Tang},
  title        = {{GPT} understands, too},
  journal      = {{AI} Open},
  volume       = {5},
  pages        = {208--215},
  year         = {2024},
  url          = {https://doi.org/10.1016/j.aiopen.2023.08.012},
  doi          = {10.1016/J.AIOPEN.2023.08.012},
  bibsource    = {dblp computer science bibliography, https://dblp.org}
}

@inproceedings{prompt-tuning,
  author       = {Brian Lester and
                  Rami Al{-}Rfou and
                  Noah Constant},
  editor       = {Marie{-}Francine Moens and
                  Xuanjing Huang and
                  Lucia Specia and
                  Scott Wen{-}tau Yih},
  title        = {The Power of Scale for Parameter-Efficient Prompt Tuning},
  booktitle    = {Proceedings of the 2021 Conference on Empirical Methods in Natural
                  Language Processing, {EMNLP} 2021, Virtual Event / Punta Cana, Dominican
                  Republic, 7-11 November, 2021},
  pages        = {3045--3059},
  publisher    = {Association for Computational Linguistics},
  year         = {2021},
  url          = {https://doi.org/10.18653/v1/2021.emnlp-main.243},
  doi          = {10.18653/V1/2021.EMNLP-MAIN.243},
  bibsource    = {dblp computer science bibliography, https://dblp.org}
}

@inproceedings{prefix-Tuning,
  author       = {Xiang Lisa Li and
                  Percy Liang},
  editor       = {Chengqing Zong and
                  Fei Xia and
                  Wenjie Li and
                  Roberto Navigli},
  title        = {Prefix-Tuning: Optimizing Continuous Prompts for Generation},
  booktitle    = {Proceedings of the 59th Annual Meeting of the Association for Computational
                  Linguistics and the 11th International Joint Conference on Natural
                  Language Processing, {ACL/IJCNLP} 2021, (Volume 1: Long Papers), Virtual
                  Event, August 1-6, 2021},
  pages        = {4582--4597},
  publisher    = {Association for Computational Linguistics},
  year         = {2021},
  url          = {https://doi.org/10.18653/v1/2021.acl-long.353},
  doi          = {10.18653/V1/2021.ACL-LONG.353},
  bibsource    = {dblp computer science bibliography, https://dblp.org}
}

@inproceedings{wild,
  author       = {Wanli Yang and
                  Fei Sun and
                  Jiajun Tan and
                  Xinyu Ma and
                  Qi Cao and
                  Dawei Yin and
                  Huawei Shen and
                  Xueqi Cheng},
  editor       = {Wanxiang Che and
                  Joyce Nabende and
                  Ekaterina Shutova and
                  Mohammad Taher Pilehvar},
  title        = {The Mirage of Model Editing: Revisiting Evaluation in the Wild},
  booktitle    = {Proceedings of the 63rd Annual Meeting of the Association for Computational
                  Linguistics (Volume 1: Long Papers), {ACL} 2025, Vienna, Austria,
                  July 27 - August 1, 2025},
  pages        = {15336--15354},
  publisher    = {Association for Computational Linguistics},
  year         = {2025},
  url          = {https://aclanthology.org/2025.acl-long.745/},
  bibsource    = {dblp computer science bibliography, https://dblp.org}
}

@inproceedings{ice,
  author       = {Siyuan Qi and
                  Bangcheng Yang and
                  Kailin Jiang and
                  Xiaobo Wang and
                  Jiaqi Li and
                  Yifan Zhong and
                  Yaodong Yang and
                  Zilong Zheng},
  title        = {In-Context Editing: Learning Knowledge from Self-Induced Distributions},
  booktitle    = {The Thirteenth International Conference on Learning Representations,
                  {ICLR} 2025, Singapore, April 24-28, 2025},
  publisher    = {OpenReview.net},
  year         = {2025},
  url          = {https://openreview.net/forum?id=w6rHCuN3YG},
  bibsource    = {dblp computer science bibliography, https://dblp.org}
}

@inproceedings{atbias,
  author       = {Baolong Bi and
                  Shenghua Liu and
                  Yiwei Wang and
                  Lingrui Mei and
                  Hongcheng Gao and
                  Yilong Xu and
                  Xueqi Cheng},
  editor       = {Yaser Al{-}Onaizan and
                  Mohit Bansal and
                  Yun{-}Nung Chen},
  title        = {Adaptive Token Biaser: Knowledge Editing via Biasing Key Entities},
  booktitle    = {Findings of the Association for Computational Linguistics: {EMNLP}
                  2024, Miami, Florida, USA, November 12-16, 2024},
  series       = {Findings of {ACL}},
  volume       = {{EMNLP} 2024},
  pages        = {11071--11083},
  publisher    = {Association for Computational Linguistics},
  year         = {2024},
  url          = {https://doi.org/10.18653/v1/2024.findings-emnlp.647},
  doi          = {10.18653/V1/2024.FINDINGS-EMNLP.647},
  bibsource    = {dblp computer science bibliography, https://dblp.org}
}

@article{dr-ike,
  author       = {Mahmud Wasif Nafee and
                  Maiqi Jiang and
                  Haipeng Chen and
                  Yanfu Zhang},
  title        = {Dynamic Retriever for In-Context Knowledge Editing via Policy Optimization},
  journal      = {CoRR},
  volume       = {abs/2510.21059},
  year         = {2025},
  url          = {https://doi.org/10.48550/arXiv.2510.21059},
  doi          = {10.48550/ARXIV.2510.21059},
  eprinttype    = {arXiv},
  eprint       = {2510.21059},
  bibsource    = {dblp computer science bibliography, https://dblp.org}
}

@inproceedings{recipe,
  author       = {Qizhou Chen and
                  Taolin Zhang and
                  Xiaofeng He and
                  Dongyang Li and
                  Chengyu Wang and
                  Longtao Huang and
                  Hui Xue'},
  editor       = {Yaser Al{-}Onaizan and
                  Mohit Bansal and
                  Yun{-}Nung Chen},
  title        = {Lifelong Knowledge Editing for LLMs with Retrieval-Augmented Continuous
                  Prompt Learning},
  booktitle    = {Proceedings of the 2024 Conference on Empirical Methods in Natural
                  Language Processing, {EMNLP} 2024, Miami, FL, USA, November 12-16,
                  2024},
  pages        = {13565--13580},
  publisher    = {Association for Computational Linguistics},
  year         = {2024},
  url          = {https://doi.org/10.18653/v1/2024.emnlp-main.751},
  doi          = {10.18653/V1/2024.EMNLP-MAIN.751},
  bibsource    = {dblp computer science bibliography, https://dblp.org}
}

@inproceedings{mmlu,
  author       = {Dan Hendrycks and
                  Collin Burns and
                  Steven Basart and
                  Andy Zou and
                  Mantas Mazeika and
                  Dawn Song and
                  Jacob Steinhardt},
  title        = {Measuring Massive Multitask Language Understanding},
  booktitle    = {9th International Conference on Learning Representations, {ICLR} 2021,
                  Virtual Event, Austria, May 3-7, 2021},
  publisher    = {OpenReview.net},
  year         = {2021},
  url          = {https://openreview.net/forum?id=d7KBjmI3GmQ},
  bibsource    = {dblp computer science bibliography, https://dblp.org}
}

@article{bge-base-en-v1.57,
  author       = {Shitao Xiao and
                  Zheng Liu and
                  Peitian Zhang and
                  Niklas Muennighoff},
  title        = {C-Pack: Packaged Resources To Advance General Chinese Embedding},
  journal      = {CoRR},
  volume       = {abs/2309.07597},
  year         = {2023},
  url          = {https://doi.org/10.48550/arXiv.2309.07597},
  doi          = {10.48550/ARXIV.2309.07597},
  eprinttype    = {arXiv},
  eprint       = {2309.07597},
  bibsource    = {dblp computer science bibliography, https://dblp.org}
}

@inproceedings{Yao,
  author       = {Yunzhi Yao and
                  Peng Wang and
                  Bozhong Tian and
                  Siyuan Cheng and
                  Zhoubo Li and
                  Shumin Deng and
                  Huajun Chen and
                  Ningyu Zhang},
  editor       = {Houda Bouamor and
                  Juan Pino and
                  Kalika Bali},
  title        = {Editing Large Language Models: Problems, Methods, and Opportunities},
  booktitle    = {Proceedings of the 2023 Conference on Empirical Methods in Natural
                  Language Processing, {EMNLP} 2023, Singapore, December 6-10, 2023},
  pages        = {10222--10240},
  publisher    = {Association for Computational Linguistics},
  year         = {2023},
  url          = {https://doi.org/10.18653/v1/2023.emnlp-main.632},
  doi          = {10.18653/V1/2023.EMNLP-MAIN.632},
  bibsource    = {dblp computer science bibliography, https://dblp.org}
}

@article{ripple-effects,
  author       = {Roi Cohen and
                  Eden Biran and
                  Ori Yoran and
                  Amir Globerson and
                  Mor Geva},
  title        = {Evaluating the Ripple Effects of Knowledge Editing in Language Models},
  journal      = {Trans. Assoc. Comput. Linguistics},
  volume       = {12},
  pages        = {283--298},
  year         = {2024},
  url          = {https://doi.org/10.1162/tacl\_a\_00644},
  doi          = {10.1162/TACL\_A\_00644},
  bibsource    = {dblp computer science bibliography, https://dblp.org}
}

@inproceedings{mquake,
  author       = {Zexuan Zhong and
                  Zhengxuan Wu and
                  Christopher D. Manning and
                  Christopher Potts and
                  Danqi Chen},
  editor       = {Houda Bouamor and
                  Juan Pino and
                  Kalika Bali},
  title        = {MQuAKE: Assessing Knowledge Editing in Language Models via Multi-Hop
                  Questions},
  booktitle    = {Proceedings of the 2023 Conference on Empirical Methods in Natural
                  Language Processing, {EMNLP} 2023, Singapore, December 6-10, 2023},
  pages        = {15686--15702},
  publisher    = {Association for Computational Linguistics},
  year         = {2023},
  url          = {https://doi.org/10.18653/v1/2023.emnlp-main.971},
  doi          = {10.18653/V1/2023.EMNLP-MAIN.971},
  bibsource    = {dblp computer science bibliography, https://dblp.org}
}

@inproceedings{dune,
  author       = {Afra Feyza Aky{\"{u}}rek and
                  Eric Pan and
                  Garry Kuwanto and
                  Derry Wijaya},
  editor       = {Houda Bouamor and
                  Juan Pino and
                  Kalika Bali},
  title        = {DUnE: Dataset for Unified Editing},
  booktitle    = {Proceedings of the 2023 Conference on Empirical Methods in Natural
                  Language Processing, {EMNLP} 2023, Singapore, December 6-10, 2023},
  pages        = {1847--1861},
  publisher    = {Association for Computational Linguistics},
  year         = {2023},
  url          = {https://doi.org/10.18653/v1/2023.emnlp-main.114},
  doi          = {10.18653/V1/2023.EMNLP-MAIN.114},
  bibsource    = {dblp computer science bibliography, https://dblp.org}
}

@inproceedings{LoRA,
  author       = {Edward J. Hu and
                  Yelong Shen and
                  Phillip Wallis and
                  Zeyuan Allen{-}Zhu and
                  Yuanzhi Li and
                  Shean Wang and
                  Lu Wang and
                  Weizhu Chen},
  title        = {LoRA: Low-Rank Adaptation of Large Language Models},
  booktitle    = {The Tenth International Conference on Learning Representations, {ICLR}
                  2022, Virtual Event, April 25-29, 2022},
  publisher    = {OpenReview.net},
  year         = {2022},
  url          = {https://openreview.net/forum?id=nZeVKeeFYf9},
  bibsource    = {dblp computer science bibliography, https://dblp.org}
}

@inproceedings{ike,
  author       = {Ce Zheng and
                  Lei Li and
                  Qingxiu Dong and
                  Yuxuan Fan and
                  Zhiyong Wu and
                  Jingjing Xu and
                  Baobao Chang},
  editor       = {Houda Bouamor and
                  Juan Pino and
                  Kalika Bali},
  title        = {Can We Edit Factual Knowledge by In-Context Learning?},
  booktitle    = {Proceedings of the 2023 Conference on Empirical Methods in Natural
                  Language Processing, {EMNLP} 2023, Singapore, December 6-10, 2023},
  pages        = {4862--4876},
  publisher    = {Association for Computational Linguistics},
  year         = {2023},
  url          = {https://aclanthology.org/2023.emnlp-main.296},
  bibsource    = {dblp computer science bibliography, https://dblp.org}
}

@inproceedings{ke,
    title = "Editing Factual Knowledge in Language Models",
    author = "De Cao, Nicola  and
      Aziz, Wilker  and
      Titov, Ivan",
    editor = "Moens, Marie-Francine  and
      Huang, Xuanjing  and
      Specia, Lucia  and
      Yih, Scott Wen-tau",
    booktitle = "Proceedings of the 2021 Conference on Empirical Methods in Natural Language Processing",
    month = nov,
    year = "2021",
    address = "Online and Punta Cana, Dominican Republic",
    publisher = "Association for Computational Linguistics",
    url = "https://aclanthology.org/2021.emnlp-main.522",
    doi = "10.18653/v1/2021.emnlp-main.522",
    pages = "6491--6506",
}

@inproceedings{coachhook,
  author       = {Shuaiyi Li and
                  Yang Deng and
                  Deng Cai and
                  Hongyuan Lu and
                  Liang Chen and
                  Wai Lam},
  editor       = {Yaser Al{-}Onaizan and
                  Mohit Bansal and
                  Yun{-}Nung Chen},
  title        = {Consecutive Batch Model Editing with HooK Layers},
  booktitle    = {Proceedings of the 2024 Conference on Empirical Methods in Natural
                  Language Processing, {EMNLP} 2024, Miami, FL, USA, November 12-16,
                  2024},
  pages        = {13817--13833},
  publisher    = {Association for Computational Linguistics},
  year         = {2024},
  url          = {https://aclanthology.org/2024.emnlp-main.765},
  bibsource    = {dblp computer science bibliography, https://dblp.org}
}

@inproceedings{alphaedit,
  author       = {Junfeng Fang and
                  Houcheng Jiang and
                  Kun Wang and
                  Yunshan Ma and
                  Jie Shi and
                  Xiang Wang and
                  Xiangnan He and
                  Tat{-}Seng Chua},
  title        = {AlphaEdit: Null-Space Constrained Knowledge Editing for Language Models},
  booktitle    = {The Thirteenth International Conference on Learning Representations,
                  {ICLR} 2025, Singapore, April 24-28, 2025},
  publisher    = {OpenReview.net},
  year         = {2025},
  url          = {https://openreview.net/forum?id=HvSytvg3Jh},
  bibsource    = {dblp computer science bibliography, https://dblp.org}
}

@inproceedings{melo,
  author       = {Lang Yu and
                  Qin Chen and
                  Jie Zhou and
                  Liang He},
  editor       = {Michael J. Wooldridge and
                  Jennifer G. Dy and
                  Sriraam Natarajan},
  title        = {{MELO:} Enhancing Model Editing with Neuron-Indexed Dynamic LoRA},
  booktitle    = {Thirty-Eighth {AAAI} Conference on Artificial Intelligence, {AAAI}
                  2024, Thirty-Sixth Conference on Innovative Applications of Artificial
                  Intelligence, {IAAI} 2024, Fourteenth Symposium on Educational Advances
                  in Artificial Intelligence, {EAAI} 2014, February 20-27, 2024, Vancouver,
                  Canada},
  pages        = {19449--19457},
  publisher    = {{AAAI} Press},
  year         = {2024},
  url          = {https://doi.org/10.1609/aaai.v38i17.29916},
  doi          = {10.1609/AAAI.V38I17.29916},
  bibsource    = {dblp computer science bibliography, https://dblp.org}
}

@inproceedings{calinet,
  author       = {Qingxiu Dong and
                  Damai Dai and
                  Yifan Song and
                  Jingjing Xu and
                  Zhifang Sui and
                  Lei Li},
  editor       = {Yoav Goldberg and
                  Zornitsa Kozareva and
                  Yue Zhang},
  title        = {Calibrating Factual Knowledge in Pretrained Language Models},
  booktitle    = {Findings of the Association for Computational Linguistics: {EMNLP}
                  2022, Abu Dhabi, United Arab Emirates, December 7-11, 2022},
  pages        = {5937--5947},
  publisher    = {Association for Computational Linguistics},
  year         = {2022},
  url          = {https://doi.org/10.18653/v1/2022.findings-emnlp.438},
  doi          = {10.18653/V1/2022.FINDINGS-EMNLP.438},
  bibsource    = {dblp computer science bibliography, https://dblp.org}
}

@article{pmet,
  author       = {Xiaopeng Li and
                  Shasha Li and
                  Shezheng Song and
                  Jing Yang and
                  Jun Ma and
                  Jie Yu},
  title        = {{PMET:} Precise Model Editing in a Transformer},
  journal      = {CoRR},
  volume       = {abs/2308.08742},
  year         = {2023},
  url          = {https://doi.org/10.48550/arXiv.2308.08742},
  doi          = {10.48550/ARXIV.2308.08742},
  eprinttype    = {arXiv},
  eprint       = {2308.08742},
  bibsource    = {dblp computer science bibliography, https://dblp.org}
}

@article{grace,
  author       = {Thomas Hartvigsen and
                  Swami Sankaranarayanan and
                  Hamid Palangi and
                  Yoon Kim and
                  Marzyeh Ghassemi},
  title        = {Aging with {GRACE:} Lifelong Model Editing with Discrete Key-Value
                  Adaptors},
  journal      = {CoRR},
  volume       = {abs/2211.11031},
  year         = {2022},
  url          = {https://doi.org/10.48550/arXiv.2211.11031},
  doi          = {10.48550/ARXIV.2211.11031},
  eprinttype    = {arXiv},
  eprint       = {2211.11031},
  bibsource    = {dblp computer science bibliography, https://dblp.org}
}

@inproceedings{rome,
  author       = {Kevin Meng and
                  David Bau and
                  Alex Andonian and
                  Yonatan Belinkov},
  editor       = {Sanmi Koyejo and
                  S. Mohamed and
                  A. Agarwal and
                  Danielle Belgrave and
                  K. Cho and
                  A. Oh},
  title        = {Locating and Editing Factual Associations in {GPT}},
  booktitle    = {Advances in Neural Information Processing Systems 35: Annual Conference
                  on Neural Information Processing Systems 2022, NeurIPS 2022, New Orleans,
                  LA, USA, November 28 - December 9, 2022},
  year         = {2022},
  url          = {http://papers.nips.cc/paper\_files/paper/2022/hash/6f1d43d5a82a37e89b0665b33bf3a182-Abstract-Conference.html},
  bibsource    = {dblp computer science bibliography, https://dblp.org}
}

@inproceedings{mend,
  author       = {Eric Mitchell and
                  Charles Lin and
                  Antoine Bosselut and
                  Chelsea Finn and
                  Christopher D. Manning},
  title        = {Fast Model Editing at Scale},
  booktitle    = {The Tenth International Conference on Learning Representations, {ICLR}
                  2022, Virtual Event, April 25-29, 2022},
  publisher    = {OpenReview.net},
  year         = {2022},
  url          = {https://openreview.net/forum?id=0DcZxeWfOPt},
  bibsource    = {dblp computer science bibliography, https://dblp.org}
}

@inproceedings{serac,
  author       = {Eric Mitchell and
                  Charles Lin and
                  Antoine Bosselut and
                  Christopher D. Manning and
                  Chelsea Finn},
  editor       = {Kamalika Chaudhuri and
                  Stefanie Jegelka and
                  Le Song and
                  Csaba Szepesv{\'{a}}ri and
                  Gang Niu and
                  Sivan Sabato},
  title        = {Memory-Based Model Editing at Scale},
  booktitle    = {International Conference on Machine Learning, {ICML} 2022, 17-23 July
                  2022, Baltimore, Maryland, {USA}},
  series       = {Proceedings of Machine Learning Research},
  volume       = {162},
  pages        = {15817--15831},
  publisher    = {{PMLR}},
  year         = {2022},
  url          = {https://proceedings.mlr.press/v162/mitchell22a.html},
  bibsource    = {dblp computer science bibliography, https://dblp.org}
}

@inproceedings{memit,
  author       = {Kevin Meng and
                  Arnab Sen Sharma and
                  Alex J. Andonian and
                  Yonatan Belinkov and
                  David Bau},
  title        = {Mass-Editing Memory in a Transformer},
  booktitle    = {The Eleventh International Conference on Learning Representations,
                  {ICLR} 2023, Kigali, Rwanda, May 1-5, 2023},
  publisher    = {OpenReview.net},
  year         = {2023},
  url          = {https://openreview.net/pdf?id=MkbcAHIYgyS},
  bibsource    = {dblp computer science bibliography, https://dblp.org}
}

@inproceedings{r-rome,
  author       = {Akshat Gupta and
                  Sidharth Baskaran and
                  Gopala Anumanchipalli},
  editor       = {Yaser Al{-}Onaizan and
                  Mohit Bansal and
                  Yun{-}Nung Chen},
  title        = {Rebuilding {ROME} : Resolving Model Collapse during Sequential Model
                  Editing},
  booktitle    = {Proceedings of the 2024 Conference on Empirical Methods in Natural
                  Language Processing, {EMNLP} 2024, Miami, FL, USA, November 12-16,
                  2024},
  pages        = {21738--21744},
  publisher    = {Association for Computational Linguistics},
  year         = {2024},
  url          = {https://aclanthology.org/2024.emnlp-main.1210},
  bibsource    = {dblp computer science bibliography, https://dblp.org}
}

@inproceedings{t-patcher,
  author       = {Zeyu Huang and
                  Yikang Shen and
                  Xiaofeng Zhang and
                  Jie Zhou and
                  Wenge Rong and
                  Zhang Xiong},
  title        = {Transformer-Patcher: One Mistake Worth One Neuron},
  booktitle    = {The Eleventh International Conference on Learning Representations,
                  {ICLR} 2023, Kigali, Rwanda, May 1-5, 2023},
  publisher    = {OpenReview.net},
  year         = {2023},
  url          = {https://openreview.net/pdf?id=4oYUGeGBPm},
  bibsource    = {dblp computer science bibliography, https://dblp.org}
}

@inproceedings{kn,
  author       = {Damai Dai and
                  Li Dong and
                  Yaru Hao and
                  Zhifang Sui and
                  Baobao Chang and
                  Furu Wei},
  editor       = {Smaranda Muresan and
                  Preslav Nakov and
                  Aline Villavicencio},
  title        = {Knowledge Neurons in Pretrained Transformers},
  booktitle    = {Proceedings of the 60th Annual Meeting of the Association for Computational
                  Linguistics (Volume 1: Long Papers), {ACL} 2022, Dublin, Ireland,
                  May 22-27, 2022},
  pages        = {8493--8502},
  publisher    = {Association for Computational Linguistics},
  year         = {2022},
  url          = {https://doi.org/10.18653/v1/2022.acl-long.581},
  doi          = {10.18653/V1/2022.ACL-LONG.581},
  bibsource    = {dblp computer science bibliography, https://dblp.org}
}

@inproceedings{wise,
  author       = {Peng Wang and
                  Zexi Li and
                  Ningyu Zhang and
                  Ziwen Xu and
                  Yunzhi Yao and
                  Yong Jiang and
                  Pengjun Xie and
                  Fei Huang and
                  Huajun Chen},
  editor       = {Amir Globersons and
                  Lester Mackey and
                  Danielle Belgrave and
                  Angela Fan and
                  Ulrich Paquet and
                  Jakub M. Tomczak and
                  Cheng Zhang},
  title        = {{WISE:} Rethinking the Knowledge Memory for Lifelong Model Editing
                  of Large Language Models},
  booktitle    = {Advances in Neural Information Processing Systems 38: Annual Conference
                  on Neural Information Processing Systems 2024, NeurIPS 2024, Vancouver,
                  BC, Canada, December 10 - 15, 2024},
  year         = {2024},
  url          = {http://papers.nips.cc/paper\_files/paper/2024/hash/60960ad78868fce5c165295fbd895060-Abstract-Conference.html},
  bibsource    = {dblp computer science bibliography, https://dblp.org}
}

@inproceedings{emmet,
  author       = {Akshat Gupta and
                  Dev Sajnani and
                  Gopala Anumanchipalli},
  editor       = {Yaser Al{-}Onaizan and
                  Mohit Bansal and
                  Yun{-}Nung Chen},
  title        = {A Unified Framework for Model Editing},
  booktitle    = {Findings of the Association for Computational Linguistics: {EMNLP}
                  2024, Miami, Florida, USA, November 12-16, 2024},
  pages        = {15403--15418},
  publisher    = {Association for Computational Linguistics},
  year         = {2024},
  url          = {https://aclanthology.org/2024.findings-emnlp.903},
  bibsource    = {dblp computer science bibliography, https://dblp.org}
}

@inproceedings{gisting,
  author       = {Jesse Mu and
                  Xiang Li and
                  Noah D. Goodman},
  editor       = {Alice Oh and
                  Tristan Naumann and
                  Amir Globerson and
                  Kate Saenko and
                  Moritz Hardt and
                  Sergey Levine},
  title        = {Learning to Compress Prompts with Gist Tokens},
  booktitle    = {Advances in Neural Information Processing Systems 36: Annual Conference
                  on Neural Information Processing Systems 2023, NeurIPS 2023, New Orleans,
                  LA, USA, December 10 - 16, 2023},
  year         = {2023},
  url          = {http://papers.nips.cc/paper\_files/paper/2023/hash/3d77c6dcc7f143aa2154e7f4d5e22d68-Abstract-Conference.html},
  bibsource    = {dblp computer science bibliography, https://dblp.org}
}

@article{distillation,
  author       = {Geoffrey E. Hinton and
                  Oriol Vinyals and
                  Jeffrey Dean},
  title        = {Distilling the Knowledge in a Neural Network},
  journal      = {CoRR},
  volume       = {abs/1503.02531},
  year         = {2015},
  url          = {http://arxiv.org/abs/1503.02531},
  eprinttype    = {arXiv},
  eprint       = {1503.02531},
  bibsource    = {dblp computer science bibliography, https://dblp.org}
}

@inproceedings{icl,
  author       = {Tom B. Brown and
                  Benjamin Mann and
                  Nick Ryder and
                  Melanie Subbiah and
                  Jared Kaplan and
                  Prafulla Dhariwal and
                  Arvind Neelakantan and
                  Pranav Shyam and
                  Girish Sastry and
                  Amanda Askell and
                  Sandhini Agarwal and
                  Ariel Herbert{-}Voss and
                  Gretchen Krueger and
                  Tom Henighan and
                  Rewon Child and
                  Aditya Ramesh and
                  Daniel M. Ziegler and
                  Jeffrey Wu and
                  Clemens Winter and
                  Christopher Hesse and
                  Mark Chen and
                  Eric Sigler and
                  Mateusz Litwin and
                  Scott Gray and
                  Benjamin Chess and
                  Jack Clark and
                  Christopher Berner and
                  Sam McCandlish and
                  Alec Radford and
                  Ilya Sutskever and
                  Dario Amodei},
  editor       = {Hugo Larochelle and
                  Marc'Aurelio Ranzato and
                  Raia Hadsell and
                  Maria{-}Florina Balcan and
                  Hsuan{-}Tien Lin},
  title        = {Language Models are Few-Shot Learners},
  booktitle    = {Advances in Neural Information Processing Systems 33: Annual Conference
                  on Neural Information Processing Systems 2020, NeurIPS 2020, December
                  6-12, 2020, virtual},
  year         = {2020},
  url          = {https://proceedings.neurips.cc/paper/2020/hash/1457c0d6bfcb4967418bfb8ac142f64a-Abstract.html},
  bibsource    = {dblp computer science bibliography, https://dblp.org}
}

@inproceedings{hurt-general-ability,
  author       = {Jia{-}Chen Gu and
                  Hao{-}Xiang Xu and
                  Jun{-}Yu Ma and
                  Pan Lu and
                  Zhen{-}Hua Ling and
                  Kai{-}Wei Chang and
                  Nanyun Peng},
  editor       = {Yaser Al{-}Onaizan and
                  Mohit Bansal and
                  Yun{-}Nung Chen},
  title        = {Model Editing Harms General Abilities of Large Language Models: Regularization
                  to the Rescue},
  booktitle    = {Proceedings of the 2024 Conference on Empirical Methods in Natural
                  Language Processing, {EMNLP} 2024, Miami, FL, USA, November 12-16,
                  2024},
  pages        = {16801--16819},
  publisher    = {Association for Computational Linguistics},
  year         = {2024},
  url          = {https://aclanthology.org/2024.emnlp-main.934},
  bibsource    = {dblp computer science bibliography, https://dblp.org}
}

@inproceedings{Ocean-with-a-Spoon,
  author       = {Yuval Pinter and
                  Michael Elhadad},
  editor       = {Houda Bouamor and
                  Juan Pino and
                  Kalika Bali},
  title        = {Emptying the Ocean with a Spoon: Should We Edit Models?},
  booktitle    = {Findings of the Association for Computational Linguistics: {EMNLP}
                  2023, Singapore, December 6-10, 2023},
  pages        = {15164--15172},
  publisher    = {Association for Computational Linguistics},
  year         = {2023},
  url          = {https://doi.org/10.18653/v1/2023.findings-emnlp.1012},
  doi          = {10.18653/V1/2023.FINDINGS-EMNLP.1012},
  bibsource    = {dblp computer science bibliography, https://dblp.org}
}

@inproceedings{zsre,
    title = "Zero-Shot Relation Extraction via Reading Comprehension",
    author = "Levy, Omer  and
      Seo, Minjoon  and
      Choi, Eunsol  and
      Zettlemoyer, Luke",
    editor = "Levy, Roger  and
      Specia, Lucia",
    booktitle = "Proceedings of the 21st Conference on Computational Natural Language Learning ({C}o{NLL} 2017)",
    month = aug,
    year = "2017",
    address = "Vancouver, Canada",
    publisher = "Association for Computational Linguistics",
    url = "https://aclanthology.org/K17-1034",
    doi = "10.18653/v1/K17-1034",
    pages = "333--342",
}

@article{knowedit,
  author       = {Ningyu Zhang and
                  Yunzhi Yao and
                  Bozhong Tian and
                  Peng Wang and
                  Shumin Deng and
                  Mengru Wang and
                  Zekun Xi and
                  Shengyu Mao and
                  Jintian Zhang and
                  Yuansheng Ni and
                  Siyuan Cheng and
                  Ziwen Xu and
                  Xin Xu and
                  Jia{-}Chen Gu and
                  Yong Jiang and
                  Pengjun Xie and
                  Fei Huang and
                  Lei Liang and
                  Zhiqiang Zhang and
                  Xiaowei Zhu and
                  Jun Zhou and
                  Huajun Chen},
  title        = {A Comprehensive Study of Knowledge Editing for Large Language Models},
  journal      = {CoRR},
  volume       = {abs/2401.01286},
  year         = {2024},
  url          = {https://doi.org/10.48550/arXiv.2401.01286},
  doi          = {10.48550/ARXIV.2401.01286},
  eprinttype    = {arXiv},
  eprint       = {2401.01286},
  bibsource    = {dblp computer science bibliography, https://dblp.org}
}

@article{atten-entropy,
  author       = {Zhisong Zhang and
                  Yan Wang and
                  Xinting Huang and
                  Tianqing Fang and
                  Hongming Zhang and
                  Chenlong Deng and
                  Shuaiyi Li and
                  Dong Yu},
  title        = {Attention Entropy is a Key Factor: An Analysis of Parallel Context
                  Encoding with Full-attention-based Pre-trained Language Models},
  journal      = {CoRR},
  volume       = {abs/2412.16545},
  year         = {2024},
  url          = {https://doi.org/10.48550/arXiv.2412.16545},
  doi          = {10.48550/ARXIV.2412.16545},
  eprinttype    = {arXiv},
  eprint       = {2412.16545},
  bibsource    = {dblp computer science bibliography, https://dblp.org}
}

@inproceedings{edit-overfit,
  author       = {Mengqi Zhang and
                  Xiaotian Ye and
                  Qiang Liu and
                  Shu Wu and
                  Pengjie Ren and
                  Zhumin Chen},
  title        = {Uncovering Overfitting in Large Language Model Editing},
  booktitle    = {The Thirteenth International Conference on Learning Representations,
                  {ICLR} 2025, Singapore, April 24-28, 2025},
  publisher    = {OpenReview.net},
  year         = {2025},
  url          = {https://openreview.net/forum?id=t8qcGXaepr},
  bibsource    = {dblp computer science bibliography, https://dblp.org}
}

@article{distil-bert,
  author       = {Victor Sanh and
                  Lysandre Debut and
                  Julien Chaumond and
                  Thomas Wolf},
  title        = {DistilBERT, a distilled version of {BERT:} smaller, faster, cheaper
                  and lighter},
  journal      = {CoRR},
  volume       = {abs/1910.01108},
  year         = {2019},
  url          = {http://arxiv.org/abs/1910.01108},
  eprinttype    = {arXiv},
  eprint       = {1910.01108},
  bibsource    = {dblp computer science bibliography, https://dblp.org}
}

@article{qwen2,
  author       = {An Yang and
                  Baosong Yang and
                  Binyuan Hui and
                  Bo Zheng and
                  Bowen Yu and
                  Chang Zhou and
                  Chengpeng Li and
                  Chengyuan Li and
                  Dayiheng Liu and
                  Fei Huang and
                  Guanting Dong and
                  Haoran Wei and
                  Huan Lin and
                  Jialong Tang and
                  Jialin Wang and
                  Jian Yang and
                  Jianhong Tu and
                  Jianwei Zhang and
                  Jianxin Ma and
                  Jianxin Yang and
                  Jin Xu and
                  Jingren Zhou and
                  Jinze Bai and
                  Jinzheng He and
                  Junyang Lin and
                  Kai Dang and
                  Keming Lu and
                  Keqin Chen and
                  Kexin Yang and
                  Mei Li and
                  Mingfeng Xue and
                  Na Ni and
                  Pei Zhang and
                  Peng Wang and
                  Ru Peng and
                  Rui Men and
                  Ruize Gao and
                  Runji Lin and
                  Shijie Wang and
                  Shuai Bai and
                  Sinan Tan and
                  Tianhang Zhu and
                  Tianhao Li and
                  Tianyu Liu and
                  Wenbin Ge and
                  Xiaodong Deng and
                  Xiaohuan Zhou and
                  Xingzhang Ren and
                  Xinyu Zhang and
                  Xipin Wei and
                  Xuancheng Ren and
                  Xuejing Liu and
                  Yang Fan and
                  Yang Yao and
                  Yichang Zhang and
                  Yu Wan and
                  Yunfei Chu and
                  Yuqiong Liu and
                  Zeyu Cui and
                  Zhenru Zhang and
                  Zhifang Guo and
                  Zhihao Fan},
  title        = {Qwen2 Technical Report},
  journal      = {CoRR},
  volume       = {abs/2407.10671},
  year         = {2024},
  url          = {https://doi.org/10.48550/arXiv.2407.10671},
  doi          = {10.48550/ARXIV.2407.10671},
  eprinttype    = {arXiv},
  eprint       = {2407.10671},
  bibsource    = {dblp computer science bibliography, https://dblp.org}
}

@article{gpt-4,
  author       = {OpenAI},
  title        = {{GPT-4} Technical Report},
  journal      = {CoRR},
  volume       = {abs/2303.08774},
  year         = {2023},
  url          = {https://doi.org/10.48550/arXiv.2303.08774},
  doi          = {10.48550/ARXIV.2303.08774},
  eprinttype    = {arXiv},
  eprint       = {2303.08774},
  bibsource    = {dblp computer science bibliography, https://dblp.org}
}

@article{deepseek-v3,
  author       = {DeepSeek{-}AI and
                  Aixin Liu and
                  Bei Feng and
                  Bing Xue and
                  Bingxuan Wang and
                  Bochao Wu and
                  Chengda Lu and
                  Chenggang Zhao and
                  Chengqi Deng and
                  Chenyu Zhang and
                  Chong Ruan and
                  Damai Dai and
                  Daya Guo and
                  Dejian Yang and
                  Deli Chen and
                  Dongjie Ji and
                  Erhang Li and
                  Fangyun Lin and
                  Fucong Dai and
                  Fuli Luo and
                  Guangbo Hao and
                  Guanting Chen and
                  Guowei Li and
                  H. Zhang and
                  Han Bao and
                  Hanwei Xu and
                  Haocheng Wang and
                  Haowei Zhang and
                  Honghui Ding and
                  Huajian Xin and
                  Huazuo Gao and
                  Hui Li and
                  Hui Qu and
                  J. L. Cai and
                  Jian Liang and
                  Jianzhong Guo and
                  Jiaqi Ni and
                  Jiashi Li and
                  Jiawei Wang and
                  Jin Chen and
                  Jingchang Chen and
                  Jingyang Yuan and
                  Junjie Qiu and
                  Junlong Li and
                  Junxiao Song and
                  Kai Dong and
                  Kai Hu and
                  Kaige Gao and
                  Kang Guan and
                  Kexin Huang and
                  Kuai Yu and
                  Lean Wang and
                  Lecong Zhang and
                  Lei Xu and
                  Leyi Xia and
                  Liang Zhao and
                  Litong Wang and
                  Liyue Zhang and
                  Meng Li and
                  Miaojun Wang and
                  Mingchuan Zhang and
                  Minghua Zhang and
                  Minghui Tang and
                  Mingming Li and
                  Ning Tian and
                  Panpan Huang and
                  Peiyi Wang and
                  Peng Zhang and
                  Qiancheng Wang and
                  Qihao Zhu and
                  Qinyu Chen and
                  Qiushi Du and
                  R. J. Chen and
                  R. L. Jin and
                  Ruiqi Ge and
                  Ruisong Zhang and
                  Ruizhe Pan and
                  Runji Wang and
                  Runxin Xu and
                  Ruoyu Zhang and
                  Ruyi Chen and
                  S. S. Li and
                  Shanghao Lu and
                  Shangyan Zhou and
                  Shanhuang Chen and
                  Shaoqing Wu and
                  Shengfeng Ye and
                  Shengfeng Ye and
                  Shirong Ma and
                  Shiyu Wang and
                  Shuang Zhou and
                  Shuiping Yu and
                  Shunfeng Zhou and
                  Shuting Pan and
                  T. Wang and
                  Tao Yun and
                  Tian Pei and
                  Tianyu Sun and
                  W. L. Xiao and
                  Wangding Zeng},
  title        = {DeepSeek-V3 Technical Report},
  journal      = {CoRR},
  volume       = {abs/2412.19437},
  year         = {2024},
  url          = {https://doi.org/10.48550/arXiv.2412.19437},
  doi          = {10.48550/ARXIV.2412.19437},
  eprinttype    = {arXiv},
  eprint       = {2412.19437},
  bibsource    = {dblp computer science bibliography, https://dblp.org}
}

@inproceedings{s2orc,
  author       = {Kyle Lo and
                  Lucy Lu Wang and
                  Mark Neumann and
                  Rodney Kinney and
                  Daniel S. Weld},
  editor       = {Dan Jurafsky and
                  Joyce Chai and
                  Natalie Schluter and
                  Joel R. Tetreault},
  title        = {{S2ORC:} The Semantic Scholar Open Research Corpus},
  booktitle    = {Proceedings of the 58th Annual Meeting of the Association for Computational
                  Linguistics, {ACL} 2020, Online, July 5-10, 2020},
  pages        = {4969--4983},
  publisher    = {Association for Computational Linguistics},
  year         = {2020},
  url          = {https://doi.org/10.18653/v1/2020.acl-main.447},
  doi          = {10.18653/V1/2020.ACL-MAIN.447},
  bibsource    = {dblp computer science bibliography, https://dblp.org}
}

@inproceedings{squad,
    title = "{SQ}u{AD}: 100,000+ Questions for Machine Comprehension of Text",
    author = "Rajpurkar, Pranav  and
      Zhang, Jian  and
      Lopyrev, Konstantin  and
      Liang, Percy",
    editor = "Su, Jian  and
      Duh, Kevin  and
      Carreras, Xavier",
    booktitle = "Proceedings of the 2016 Conference on Empirical Methods in Natural Language Processing",
    month = nov,
    year = "2016",
    address = "Austin, Texas",
    publisher = "Association for Computational Linguistics",
    url = "https://aclanthology.org/D16-1264",
    doi = "10.18653/v1/D16-1264",
    pages = "2383--2392",
    eprint={1606.05250},
    archivePrefix={arXiv},
    primaryClass={cs.CL},
}

@article{natural-q,
  author       = {Tom Kwiatkowski and
                  Jennimaria Palomaki and
                  Olivia Redfield and
                  Michael Collins and
                  Ankur P. Parikh and
                  Chris Alberti and
                  Danielle Epstein and
                  Illia Polosukhin and
                  Jacob Devlin and
                  Kenton Lee and
                  Kristina Toutanova and
                  Llion Jones and
                  Matthew Kelcey and
                  Ming{-}Wei Chang and
                  Andrew M. Dai and
                  Jakob Uszkoreit and
                  Quoc Le and
                  Slav Petrov},
  title        = {Natural Questions: a Benchmark for Question Answering Research},
  journal      = {Trans. Assoc. Comput. Linguistics},
  volume       = {7},
  pages        = {452--466},
  year         = {2019},
  url          = {https://doi.org/10.1162/tacl\_a\_00276},
  doi          = {10.1162/TACL\_A\_00276},
  bibsource    = {dblp computer science bibliography, https://dblp.org}
}

@inproceedings{openbookqa,
  author       = {Todor Mihaylov and
                  Peter Clark and
                  Tushar Khot and
                  Ashish Sabharwal},
  editor       = {Ellen Riloff and
                  David Chiang and
                  Julia Hockenmaier and
                  Jun'ichi Tsujii},
  title        = {Can a Suit of Armor Conduct Electricity? {A} New Dataset for Open
                  Book Question Answering},
  booktitle    = {Proceedings of the 2018 Conference on Empirical Methods in Natural
                  Language Processing, Brussels, Belgium, October 31 - November 4, 2018},
  pages        = {2381--2391},
  publisher    = {Association for Computational Linguistics},
  year         = {2018},
  url          = {https://doi.org/10.18653/v1/d18-1260},
  doi          = {10.18653/V1/D18-1260},
  bibsource    = {dblp computer science bibliography, https://dblp.org}
}

@inproceedings{qasc,
  author       = {Tushar Khot and
                  Peter Clark and
                  Michal Guerquin and
                  Peter Jansen and
                  Ashish Sabharwal},
  title        = {{QASC:} {A} Dataset for Question Answering via Sentence Composition},
  booktitle    = {The Thirty-Fourth {AAAI} Conference on Artificial Intelligence, {AAAI}
                  2020, The Thirty-Second Innovative Applications of Artificial Intelligence
                  Conference, {IAAI} 2020, The Tenth {AAAI} Symposium on Educational
                  Advances in Artificial Intelligence, {EAAI} 2020, New York, NY, USA,
                  February 7-12, 2020},
  pages        = {8082--8090},
  publisher    = {{AAAI} Press},
  year         = {2020},
  url          = {https://doi.org/10.1609/aaai.v34i05.6319},
  doi          = {10.1609/AAAI.V34I05.6319},
  bibsource    = {dblp computer science bibliography, https://dblp.org}
}

@inproceedings{medmcqa,
  author       = {Ankit Pal and
                  Logesh Kumar Umapathi and
                  Malaikannan Sankarasubbu},
  editor       = {Gerardo Flores and
                  George H. Chen and
                  Tom J. Pollard and
                  Joyce C. Ho and
                  Tristan Naumann},
  title        = {MedMCQA: {A} Large-scale Multi-Subject Multi-Choice Dataset for Medical
                  domain Question Answering},
  booktitle    = {Conference on Health, Inference, and Learning, {CHIL} 2022, 7-8 April
                  2022, Virtual Event},
  series       = {Proceedings of Machine Learning Research},
  volume       = {174},
  pages        = {248--260},
  publisher    = {{PMLR}},
  year         = {2022},
  url          = {https://proceedings.mlr.press/v174/pal22a.html},
  bibsource    = {dblp computer science bibliography, https://dblp.org}
}

@article{neteval,
  author       = {Yukai Miao and
                  Yu Bai and
                  Li Chen and
                  Dan Li and
                  Haifeng Sun and
                  Xizheng Wang and
                  Ziqiu Luo and
                  Yanyu Ren and
                  Dapeng Sun and
                  Xiuting Xu and
                  Qi Zhang and
                  Chao Xiang and
                  Xinchi Li},
  title        = {An Empirical Study of NetOps Capability of Pre-Trained Large Language
                  Models},
  journal      = {CoRR},
  volume       = {abs/2309.05557},
  year         = {2023},
  url          = {https://doi.org/10.48550/arXiv.2309.05557},
  doi          = {10.48550/ARXIV.2309.05557},
  eprinttype    = {arXiv},
  eprint       = {2309.05557},
  bibsource    = {dblp computer science bibliography, https://dblp.org}
}

@article{EasyEdit,
  author       = {Peng Wang and
                  Ningyu Zhang and
                  Bozhong Tian and
                  Zekun Xi and
                  Yunzhi Yao and
                  Ziwen Xu and
                  Mengru Wang and
                  Shengyu Mao and
                  Xiaohan Wang and
                  Siyuan Cheng and
                  Kangwei Liu and
                  Yuansheng Ni and
                  Guozhou Zheng and
                  Huajun Chen},
  title        = {EasyEdit: An Easy-to-use Knowledge Editing Framework for Large Language
                  Models},
  journal      = {CoRR},
  volume       = {abs/2308.07269},
  year         = {2023},
  url          = {https://doi.org/10.48550/arXiv.2308.07269},
  doi          = {10.48550/ARXIV.2308.07269},
  eprinttype    = {arXiv},
  eprint       = {2308.07269},
  bibsource    = {dblp computer science bibliography, https://dblp.org}
}

@inproceedings{zero,
  author       = {Samyam Rajbhandari and
                  Jeff Rasley and
                  Olatunji Ruwase and
                  Yuxiong He},
  editor       = {Christine Cuicchi and
                  Irene Qualters and
                  William T. Kramer},
  title        = {ZeRO: memory optimizations toward training trillion parameter models},
  booktitle    = {Proceedings of the International Conference for High Performance Computing,
                  Networking, Storage and Analysis, {SC} 2020, Virtual Event / Atlanta,
                  Georgia, USA, November 9-19, 2020},
  pages        = {20},
  publisher    = {{IEEE/ACM}},
  year         = {2020},
  url          = {https://doi.org/10.1109/SC41405.2020.00024},
  doi          = {10.1109/SC41405.2020.00024},
  bibsource    = {dblp computer science bibliography, https://dblp.org}
}

@inproceedings{zero-offload,
  author       = {Jie Ren and
                  Samyam Rajbhandari and
                  Reza Yazdani Aminabadi and
                  Olatunji Ruwase and
                  Shuangyan Yang and
                  Minjia Zhang and
                  Dong Li and
                  Yuxiong He},
  editor       = {Irina Calciu and
                  Geoff Kuenning},
  title        = {ZeRO-Offload: Democratizing Billion-Scale Model Training},
  booktitle    = {Proceedings of the 2021 {USENIX} Annual Technical Conference, {USENIX}
                  {ATC} 2021, July 14-16, 2021},
  pages        = {551--564},
  publisher    = {{USENIX} Association},
  year         = {2021},
  url          = {https://www.usenix.org/conference/atc21/presentation/ren-jie},
  bibsource    = {dblp computer science bibliography, https://dblp.org}
}

@inproceedings{zerO-infinity,
  author       = {Samyam Rajbhandari and
                  Olatunji Ruwase and
                  Jeff Rasley and
                  Shaden Smith and
                  Yuxiong He},
  editor       = {Bronis R. de Supinski and
                  Mary W. Hall and
                  Todd Gamblin},
  title        = {ZeRO-infinity: breaking the {GPU} memory wall for extreme scale deep
                  learning},
  booktitle    = {International Conference for High Performance Computing, Networking,
                  Storage and Analysis, {SC} 2021, St. Louis, Missouri, USA, November
                  14-19, 2021},
  pages        = {59},
  publisher    = {{ACM}},
  year         = {2021},
  url          = {https://doi.org/10.1145/3458817.3476205},
  doi          = {10.1145/3458817.3476205},
  bibsource    = {dblp computer science bibliography, https://dblp.org}
}

@article{liger-kernel,
  author       = {Pin{-}Lun Hsu and
                  Yun Dai and
                  Vignesh Kothapalli and
                  Qingquan Song and
                  Shao Tang and
                  Siyu Zhu and
                  Steven Shimizu and
                  Shivam Sahni and
                  Haowen Ning and
                  Yanning Chen},
  title        = {Liger Kernel: Efficient Triton Kernels for {LLM} Training},
  journal      = {CoRR},
  volume       = {abs/2410.10989},
  year         = {2024},
  url          = {https://doi.org/10.48550/arXiv.2410.10989},
  doi          = {10.48550/ARXIV.2410.10989},
  eprinttype    = {arXiv},
  eprint       = {2410.10989},
  bibsource    = {dblp computer science bibliography, https://dblp.org}
}

@article{unigist,
  author       = {Chenlong Deng and
                  Zhisong Zhang and
                  Kelong Mao and
                  Shuaiyi Li and
                  Tianqing Fang and
                  Hongming Zhang and
                  Haitao Mi and
                  Dong Yu and
                  Zhicheng Dou},
  title        = {UniGist: Towards General and Hardware-aligned Sequence-level Long
                  Context Compression},
  journal      = {CoRR},
  volume       = {abs/2509.15763},
  year         = {2025},
  url          = {https://doi.org/10.48550/arXiv.2509.15763},
  doi          = {10.48550/ARXIV.2509.15763},
  eprinttype    = {arXiv},
  eprint       = {2509.15763},
  bibsource    = {dblp computer science bibliography, https://dblp.org}
}

@article{silver-bullet,
  author       = {Chenlong Deng and
                  Zhisong Zhang and
                  Kelong Mao and
                  Shuaiyi Li and
                  Xinting Huang and
                  Dong Yu and
                  Zhicheng Dou},
  title        = {A Silver Bullet or a Compromise for Full Attention? {A} Comprehensive
                  Study of Gist Token-based Context Compression},
  journal      = {CoRR},
  volume       = {abs/2412.17483},
  year         = {2024},
  url          = {https://doi.org/10.48550/arXiv.2412.17483},
  doi          = {10.48550/ARXIV.2412.17483},
  eprinttype    = {arXiv},
  eprint       = {2412.17483},
  bibsource    = {dblp computer science bibliography, https://dblp.org}
}

\appendix

\section{Training details}
\label{appendix.training}
InComeS is trained on around 1.5 billion tokens, which mainly come from summarization and QA datasets. Specifically, for summerization datasets, we select $4.5e^{6}$ instances from S2ORC \cite{s2orc}, $1.15e^{6}$ instances from AG News Corpus \footnote{https://huggingface.co/datasets/sentence-transformers/agnews}; and for QA datasets, we use squad \cite{squad}, a modified version \footnote{https://huggingface.co/datasets/LLukas22/nq-simplified} of the natural question dataset \cite{natural-q}, OpenBookQA \cite{openbookqa}, QASC \cite{qasc}, MedMCQA \cite{medmcqa}, and NetEval \cite{neteval}. We also include the training split of ZsRE \cite{zsre}, COUNTERFACT \cite{memit}, $Wiki_{counterfact}$ from EasyEdit framework \cite{EasyEdit}. 

We use a cosine linear-warmup scheduler for both models, with a maximum learning rate $1e^{-5}$ and a minimum learning rate $1e^{-6}$ for Llama-3.2-1B and a maximum $5e^{-6}$ and a minimum $1e^{-6}$ for Qwen2.5-7B. To improve the model's robustness and sample them at a predefined rate during training, the batch size is dynamically sampled from a predefined set rather than a fixed number. Specifically, the predefined set for batch size is 8, 16, 32, 64, and 128, and their corresponding sample rates are 0.05, 0.05, 0.05, 0.15, and 0.7. We adopt DeepSpeed \cite{zero,zerO-infinity,zero-offload} and Liger Kernel \cite{liger-kernel} with 8 Nvidia %A100-SXM4-40G 
GPUs for distributed training. Overall, the training takes around 11 hours for Llama-3.2-1B and 35 hours for Qwen2.5-7B.

\section{Experiment details}

\begin{table*}[t]
\footnotesize
    \centering
    \setlength{\tabcolsep}{1mm}{
    \begin{adjustbox}{max width=\textwidth}
    \begin{tabular}{lccccccc}
    \toprule
    \multirow{2}{*}{Method} & \multirow{2}{*}{Model} & \multicolumn{3}{c}{$\text{WikiData}_{counterfact}$} & \multicolumn{3}{c}{ZsRE-extended} \\ 
    \cmidrule(lr){3-5}\cmidrule(lr){6-8}
    
     %\cmidrule(lr){2-4}\cmidrule(lr){5-7}\cmidrule(lr){8-10}\cmidrule(lr){11-13}
     & & Edit Success & Portability & Locality & Edit Success & Portability & Locality \\
     \midrule
     Base & \multirow{14}{*}{Llama-3.2-1B} & 21.28 / 21.28 & 19.73 / 19.73 & - & 30.06 / 30.06 & 40.17 / 40.17 & - \\
     FT-M             && 97.02 / 94.58 & 53.43 / 47.51 & 49.01 / 20.54 & 99.81 / 95.94 & 62.80 / 54.84 & 75.59 / 59.28 \\
     LoRA \cite{LoRA} && 98.91 / 82.61 & 52.87 / 43.84 & 23.31 / 14.94 & 99.86 / 93.18 & 57.43 / 44.85 & 34.94 / 37.41 \\
     ROME \cite{rome} && 94.33 / \phantom{44}-\phantom{33} & 40.44 / \phantom{44}-\phantom{33} & 26.13 / \phantom{44}-\phantom{33} & 95.41 / \phantom{44}-\phantom{33} & 46.04 / \phantom{44}-\phantom{33} & 40.19 / \phantom{44}-\phantom{33} \\
     R-ROME \cite{r-rome} && 94.31 / \phantom{44}-\phantom{33} & 40.65 / \phantom{44}-\phantom{33} & 25.83 / \phantom{44}-\phantom{33} & 95.29 / \phantom{44}-\phantom{33} & 46.46 / \phantom{44}-\phantom{33} & 39.95 / \phantom{44}-\phantom{33} \\
     MEMIT \cite{memit} && 77.87 / 66.94 & 46.34 / 23.51 & 33.45 / 11.12 & 79.43 / 58.79 & 43.26 / 25.68 & 40.33 / 30.10 \\
     EMMET \cite{emmet} && 34.56 / 27.02 & 15.43 / 08.86 & 100.0 / 100.0 & 23.68 / 15.96 & 17.54 / 06.84 & 100.0 / 100.0 \\
     MEND \cite{mend} && 38.45 / 26.66 & 24.98 / 21.06 & 17.34 / 13.54 & 53.45 / 43.33 & 37.53 / 30.77 & 46.31 / 38.75 \\
     GRACE \cite{grace} && 33.27 / 25.06 & 14.33 / 10.51 & 28.71 / 11.43 & 32.00 / 24.44 & 12.73 / 10.91 & 24.12 / 10.12 \\
     KN \cite{kn} && 20.60 / \phantom{34}-\phantom{33} & 17.16 / \phantom{34}-\phantom{33} & 19.46 / \phantom{43}-\phantom{33} & 16.01 / \phantom{43}-\phantom{33} & 06.70 / \phantom{43}-\phantom{33} & 21.23 / \phantom{34}-\phantom{33} \\
     IKE \cite{ike} && 61.70 / \phantom{34}-\phantom{33} & 45.55 / \phantom{34}-\phantom{33} & 48.80 / \phantom{43}-\phantom{33} & 59.15 / \phantom{43}-\phantom{33} & 57.39 / \phantom{43}-\phantom{33} & 40.21 / \phantom{34}-\phantom{33} \\
     SERAC \cite{serac} && 89.56 / 78.32 & 60.56 / 46.45 & 45.67 / 39.36 & 92.69 / 89.61 & 66.59 / 63.60 & 48.96 / 41.32 \\
     ICL && 93.31 / 82.95 & 65.81 / 44.75 & 52.70 / 45.59 & 68.86 / 60.84 & 62.19 / 51.58 & 62.78 / 59.43 \\
     \textbf{InComeS} && 91.16 / 76.81 & 65.15 / 45.66 & 55.22 / 53.97 & 97.22 / 87.09 & 70.70 / 52.23 & 61.33 / 59.66 \\
     \midrule
     Base & \multirow{14}{*}{Qwen2.5-7B} & 22.35 / 22.35 & 21.46 / 21.46 & - & 36.21 / 36.21 & 43.86 / 43.86 & - \\
     FT-M             && 98.93 / 90.18 & 49.39 / 43.13 & 15.73 / 12.93 & 99.51 / 92.60 & 50.04 / 46.41 & 18.00 / 27.27 \\
     LoRA \cite{LoRA} && 77.31 / 72.22 & 37.04 / 31.65 & 0.380 / 02.24 & 86.88 / 77.78 & 28.61 / 24.13 & 1.290 / 0.330 \\
     ROME \cite{rome} && 92.69 / \phantom{34}-\phantom{33} & 40.25 / \phantom{34}-\phantom{33} & 38.76 / \phantom{34}-\phantom{33} & 97.86 / \phantom{34}-\phantom{33} & 50.43 / \phantom{34}-\phantom{33} & 51.38 / \phantom{34}-\phantom{33} \\
     R-ROME \cite{r-rome} && 92.59 / \phantom{34}-\phantom{33} & 40.15 / \phantom{34}-\phantom{33} & 38.70 / \phantom{34}-\phantom{33} & 97.89 / \phantom{34}-\phantom{33} & 50.47 / \phantom{34}-\phantom{33} & 51.31 / \phantom{34}-\phantom{33} \\
     MEMIT \cite{memit} && 93.43 / 91.16 & 43.58 / 39.85 & 33.79 / 25.85 & 95.23 / 93.28 & 53.23 / 49.97 & 54.23 / 51.85 \\
     EMMET \cite{emmet} && 92.33 / 88.05 & 43.24 / 39.37 & 100.0 / 100.0 & 94.32 / 93.64 & 51.55 / 48.44 & 100.0 / 100.0 \\
     MEND \cite{mend} && 46.73 / 35.13 & 20.45 / 15.29 & 13.34 / 07.29 & 54.32 / 50.91 & 43.26 / 38.83 & 54.39 / 47.12 \\
     GRACE \cite{grace} && 31.34 / 33.77 & 25.60 / 18.55 & 19.19 / 07.29 & 33.27 / 26.79 & 14.35 / 11.25 & 12.31 / 23.45 \\
     KN \cite{kn} && 36.33 / \phantom{34}-\phantom{33} & 29.60 / \phantom{34}-\phantom{33} & 32.79 / \phantom{34}-\phantom{33} & 14.49 / \phantom{34}-\phantom{33} & 8.49 / \phantom{34}-\phantom{33} & 33.21 / \phantom{34}-\phantom{33} \\
     IKE \cite{ike} && 96.40 / \phantom{34}-\phantom{33} & 65.33 / \phantom{34}-\phantom{33} & 46.98 / \phantom{34}-\phantom{33} & 99.75 / \phantom{34}-\phantom{33} & 70.17 / \phantom{34}-\phantom{33} & 43.19 / \phantom{34}-\phantom{33} \\
     SERAC \cite{serac} && 91.79 / 80.68 & 51.12 / 41.26 & 37.84 / 35.10 & 91.12 / 82.56 & 62.41 / 52.63 & 40.56 / 42.46 \\
     ICL && 90.24 / 85.28 & 66.99 / 51.66 & 52.63 / 40.97 & 71.75 / 71.57 & 66.10 / 60.57 & 51.43 / 47.12 \\
     \textbf{InComeS} && 90.96 / 71.44 & 66.69 / 47.93 & 52.04 / 42.02 & 97.95 / 91.29 & 75.63 / 61.22 & 57.16 / 59.21 \\
     \midrule
    \bottomrule
    \end{tabular}
    \end{adjustbox}}
    \caption{More results for $WikiData_{counterfact}$ \cite{ripple-effects, knowedit} and ZsRE-extended \cite{Yao, knowedit}. The data format of each cell is in "single-edit result / 100-edits result".}
    \label{tab:portability-result-appendix}
\end{table*}

\begin{table*}[t]
\footnotesize
    \centering
    \setlength{\tabcolsep}{1.5mm}{
    \begin{adjustbox}{max width=\textwidth}
    \begin{tabular}{lcccccccc}
    \toprule
    \multirow{4}{*}{Method} & \multicolumn{6}{c}{MQuAKE} \\ 
    \cmidrule(lr){2-9}
    & \multicolumn{4}{c}{Single editing} & \multicolumn{4}{c}{Batch editing} \\ 
    \cmidrule(lr){2-5}\cmidrule(lr){6-9}
     & 2-edits & 3-edits & 4-edits & Avg & 2-edits & 3-edits & 4-edits & Avg \\
     \midrule
     \multicolumn{9}{c}{Llama-3.2-1B} \\
     \midrule
     \textbf{InComeS} & \textbf{71.19} & \textbf{72.17} & 72.62 & 71.99 & 53.93 & 52.79 & \textbf{52.73} & \textbf{53.15} \\
     - w/o zero-gist & 62.00 & 61.93 & 53.01 & 58.98 & 47.48 & 41.71 & 32.57 & 40.59 \\
     - w/ full model & 69.74 & 70.02 & \textbf{77.43} & \textbf{72.40} & 51.91 & 52.59 & 52.53 & 52.34 \\
     - w/o loss on query & 60.92 & 60.97 & 57.70 & 59.86 & 49.63 & 46.44 & 38.28 & 44.78 \\
     - w/ golden loss & 69.91 & 69.98 & 73.37 & 71.09 & \textbf{55.39} & 50.24 & 40.06 & 48.56 \\
     - w/o kl loss (Eq. \ref{eq.kl loss}) & 57.75 & 57.38 & 50.18 & 55.10 & 52.47 & 50.86 & 51.79 & 51.71 \\
     - w/o token weighting (Eq. \ref{eq.token weights}) & 57.45 & 57.84 & 49.76 & 55.02 & 53.49 & \textbf{53.20} & 51.91 & 52.63 \\
     \midrule
    \bottomrule
    \end{tabular}
    \end{adjustbox}}
    \caption{Full results of the ablation and analysis experiments.}
    \label{tab:ablation-analysis-appendix}
\end{table*}

\begin{table*}[t]
\footnotesize
    \centering
    \setlength{\tabcolsep}{1.5mm}{
    \begin{adjustbox}{max width=\textwidth}
    \begin{tabular}{lcccccccc}
    \toprule
    \multirow{2}{*}{Method} & \multicolumn{4}{c}{Single Editing} & \multicolumn{4}{c}{Batch Editing} \\ 
    \cmidrule(lr){2-5}\cmidrule(lr){6-9}
    
     %\cmidrule(lr){2-4}\cmidrule(lr){5-7}\cmidrule(lr){8-10}\cmidrule(lr){11-13}
     & 2-edits & 3-edits & 4-edits & Avg & 2-edits & 3-edits & 4-edits & Avg \\
     \midrule
     \multicolumn{9}{c}{Llama-3.2-1B} \\
     \midrule
     Base & 41.79 & 43.51 & 31.58 & 38.96 & 41.79 & 43.51 & 31.58 & 38.96\\
     FT-M             & 55.32 & 56.59 & 52.22 & 54.71 & 51.89 & 50.15 & 44.1 & 48.71 \\
     LoRA \cite{LoRA} & 67.63 & 68.84 & 60.21 & 65.56 & 50.29 & 47.88 & 47.15 & 48.44 \\
     ROME \cite{rome} & 1.97 & 7.77 & 0.55 & 3.43 & - & - & - & - \\
     R-ROME \cite{r-rome} & 2.72 & 4.29 & 8.53 & 5.18 & - & - & - & - \\
     MEMIT \cite{memit} & 40.21 & 39.54 & 22.34 & 34.03 & 29.71 & 28.87 & 15.7 & 24.76 \\
     EMMET \cite{emmet} & 5.4 & 6.61 & 4.74 & 5.58 & 11.49 & 17.16 & 15.89 & 14.85 \\
     GRACE \cite{grace} & 8.45 & 7.13 & 5.24 & 6.94 & 2.03 & 2.68 & 2.06 & 2.26 \\
     SERAC \cite{serac} & 41.84 & 43.51 & 31.58 & 38.97 & 41.84 & 43.45 & 31.74 & 39.01 \\
     MEND \cite{mend} & 39.88 & 35.23 & 33.31 & 36.41 & 35.55 & 30.29 & 29.89 & 31.91 \\
     RECIPE \cite{recipe} & 58.44 & 54.69 & 53.33 & 55.49 & 50.71 & 47.34 & 39.39 & 45.81 \\
     MeLLo \cite{mquake} & 55.21 & 54.19 & 49.87 & 53.09 & 43.33 & 47.22 & 41.76 & 44.10 \\
     MELO \cite{melo} & 44.55 & 43.99 & 32.44 & 40.33 & 43.33 & 47.22 & 33.87 & 41.47 \\
     ATBIAS \cite{atbias} & 60.35 & 61.11 & 59.46 & 60.31 & 52.41 & 51.77 & 47.86 & 50.68 \\ 
     DR-IKE \cite{dr-ike} & 52.95 & 48.26 & 47.12 & 49.44 & 42.45 & 41.33 & 32.22 & 38.67 \\
     AlphaEdit \cite{alphaedit} & 57.76 & 48.16 & 49.87 & 51.93 & 45.32 & 41.29 & 39.32 & 41.98 \\
     ICE \cite{ice} & 55.34 & 53.23 & 49.86 & 52.81 & 46.56 & 45.26 & 44.33 & 45.38 \\
     ICL & 59.23 & 59.00 & 51.63 & 56.62 & 50.06 & 49.86 & 42.37 & 47.43 \\
     \textbf{InComeS} & \textbf{71.19} & \textbf{72.17} & \textbf{72.62} & \textbf{71.99} & \textbf{53.93} & \textbf{52.79} & \textbf{52.73} & \textbf{53.15} \\
     \midrule
     \multicolumn{9}{c}{Qwen2.5-7B} \\
     \midrule
     Base & 44.08 & 44.14 & 30.62 & 39.61 & 44.08 & 44.14 & 30.62 & 39.61 \\
     FT-M & 69.89 & 74.88 & 74.96 & 73.24 & 50.31 & 49.04 & 50.39 & 49.91 \\
     LoRA \cite{LoRA} & 36.95 & 34.28 & 29.02 & 33.42 & 16.41 & 24.22 & 23.12 & 21.25 \\
     ROME \cite{rome} & 9.67 & 7.33 & 8.93 & 8.64 & - & - & - & - \\
     R-ROME \cite{r-rome} & 10.73 & 6.68 & 3.62 & 7.01 & - & - & - & - \\
     MEMIT \cite{memit} & 44.14 & 46.05 & 32.15 & 40.78 & 43.94 & 46.10 & 30.55 & 40.20 \\
     EMMET \cite{emmet} & 26.10 & 38.38 & 26.36 & 30.28 & 40.83 & 45.02 & 33.89 & 39.91 \\
     GRACE \cite{grace} & 15.78 & 13.45 & 13.23 & 14.15 & 5.43 & 7.88 & 4.23 & 5.85 \\
     SERAC \cite{serac} & 55.56 & 59.23 & 53.67 & 56.15 & 42.34 & 40.33 & 39.39 & 40.69 \\
     MEND \cite{mend} & 34.23 & 45.34 & 30.25 & 36.61 & 39.88 & 35.45 & 34.21 & 36.51 \\
     RECIPE \cite{recipe} & 57.54 & 59.69 & 58.43 & 58.55 & 49.24 & 49.78 & 40.71 & 46.58 \\
     MeLLo \cite{mquake} & 59.34 & 56.41 & 50.74 & 55.50 & 50.85 & 47.32 & 43.85 & 47.34 \\
     MELO \cite{melo} & 45.22 & 45.99 & 30.27 & 40.49 & 46.87 & 45.33 & 34.39 & 42.20 \\
     ATBIAS \cite{atbias} & 59.85 & 54.21 & 57.46 & 51.17 & 48.44 & 41.99 & 44.58 & 45.00 \\ 
     DR-IKE \cite{dr-ike} & 60.95 & 54.66 & 51.22 & 55.61 & 44.35 & 39.99 & 39.77 & 41.37 \\
     AlphaEdit \cite{alphaedit} & 57.38 & 52.76 & 50.99 & 53.71 & 56.77 & 49.99 & 44.32 & 50.36 \\
     ICE \cite{ice} & 63.23 & 67.76 & 70.91 & 67.30 & 51.33 & 47.86 & 48.33 & 49.17 \\
     ICL & \textbf{69.76} & \textbf{76.91} & 74.54 & \textbf{73.74} & 53.53 & 50.54 & 44.77 & 49.61 \\
     \textbf{InComeS} & 66.46 & 71.24 & \textbf{76.54} & 71.41 & \textbf{55.13} & \textbf{53.48} & \textbf{47.91} & \textbf{52.17} \\
     \midrule
    \bottomrule
    \end{tabular}
    \end{adjustbox}}
    \caption{Full results on MQuAKE \cite{mquake}.}
    \label{tab:mquake-appendix}
\end{table*}

\begin{table*}[t]
\footnotesize
    \centering
    \setlength{\tabcolsep}{0.5mm}{
    \begin{adjustbox}{max width=\textwidth}
    \begin{tabular}{lcccccccc}
    \toprule
    \multirow{2}{*}{Method} & \multicolumn{4}{c}{Single Editing} & \multicolumn{4}{c}{Batch Editing} \\ 
    \cmidrule(lr){2-5}\cmidrule(lr){6-9}
    
     & New info & Scientific R. & Debiasing & Avg & New info & Scientific R. & Debiasing & Avg \\
     \midrule
     \multicolumn{9}{c}{Llama-3.2-1B} \\
     \midrule
     Base & 56.85 & 55.87 & 32.73 & 48.48 & 56.85 & 55.87 & 32.73 & 48.48 \\
     FT-M & 57.43 & 53.34 & 35.73 & 48.83 & 57.07 & 53.45 & 33.43 & 47.89 \\
     LoRA \cite{LoRA} & 53.65 & 50.87 & 36.73 & 47.08 & 56.77 & 54.66 & 35.83 & 49.08 \\
     SERAC \cite{serac} & 52.67 & 48.96 & 46.73 & 49.45 & 50.76 & 47.32 & 36.52 & 44.87 \\
     ICE \cite{ice} & 54.52 & 51.32 & 39.64 & 48.49 & 56.32 & 53.44 & 36.52 & 48.76 \\
     ICL & 58.67 & 55.84 & \textbf{56.65} & 57.05 & 56.94 & 55.68 & 39.46 & 50.69 \\
     \textbf{InComeS} & \textbf{60.00} & \textbf{58.17} & 54.61 & \textbf{57.59} & \textbf{57.76} & \textbf{56.46} & \textbf{46.14} & \textbf{53.45} \\
     \midrule
     \multicolumn{9}{c}{Qwen2.5-7B} \\
     \midrule
     Base & 63.44 & 66.03 & 36.95 & 55.47 & 63.44 & 66.03 & 36.95 & 55.47 \\
     FT-M & 64.83 & 67.77 & 48.73 & 60.44 & 64.27 & 64.56 & 41.51 & 56.78 \\
     LoRA \cite{LoRA} & 63.85 & 63.22 & 39.56 & 55.54 & 62.58 & 65.33 & 43.73 & 57.21 \\
     SERAC \cite{serac} & 64.45 & 63.57 & 33.38 & 53.80 & 56.78 & 58.97 & 31.24 & 48.99 \\
     ICE \cite{ice} & 63.88 & 62.26 & 39.77 & 55.30 & 53.56 & 48.55 & 37.86 & 46.66 \\
     ICL & 65.81 & 65.34 & 35.25 & 55.47 & \textbf{66.29} & 66.24 & 40.73 & 57.75 \\
     \textbf{InComeS} & \textbf{66.83} & \textbf{68.02} & \textbf{62.59} & \textbf{65.81} & 65.61 & \textbf{67.82} & \textbf{56.69} & \textbf{63.37} \\
     \midrule
    \bottomrule
    \end{tabular}
    \end{adjustbox}}
    \caption{Full results on DUNE \cite{dune}.}
    \label{tab:dune-appendix}
\end{table*}

\subsection{Datasets}
\label{appendix.ds}

\paragraph{MQuAKE} 
The dataset MQuAKE \cite{mquake} (Multi-hop Question Answering for Knowledge Editing) is constructed based on Wikidata and contains question answering instances that require 2-hop, 3-hop, and 4-hop reasoning. In the experiment, we use the latest version of the dataset \footnote{"MQuAKE-CF-3k-v2.json" in https://github.com/princeton-nlp/MQuAKE}, which fixes the knowledge conflict problem for the old version multi-edit subset, and report the accuracy for each query.

\paragraph{DUNE}
DUNE \cite{dune} is a benchmark designed for edits in natural language form. It evaluates the model's capability of conducting natural language edits through four aspects: scientific reasoning, arithmetic reasoning, new information, and debiasing. As illustrated in Table 2 of \cite{dune}, the arithmetic reasoning edits do not follow natural language form as other subsets do and cannot represent a complete piece of instruction, therefore, we do not include it in our experiment. %In the experiment, we evaluate our method on subsets containing natural language form context (edits), which are scientific reasoning, new information, and debiasing. 

\paragraph{WikiData-counterfact}
The $\text{WikiData}_{counterfact}$ \cite{ripple-effects, knowedit} collects triplets from top-viewed pages from Wikipedia and contains portability (ripple-effect \cite{ripple-effects}) instances to test whether the output to the input relevant to the edits is changed as well. Specifically, the portability evaluates the post-edited model from three aspects, including logical generalization, subject aliasing, and reasoning.

\paragraph{ZsRE-extended}
The extended version of ZsRE \cite{knowedit, Yao} is constructed based on the original ZsRE \cite{zsre}, which is a dataset that focuses on the QA task. The extended version introduces a portability test \cite{Yao}, including inverse relation, one-hop reasoning, and subject aliasing.

\paragraph{COUNTERFACT}
COUNTERFACT \cite{memit} is a dataset that concentrates on counterfactual information, which typically receives a lower prediction score than accurate facts. It constructs out-of-scope data by substituting the subject entity with a comparable description that has the same predicate.

\subsection{Evaluation metrics}
\label{appendix.eval-metric}
This section explains the evaluation metrics used in the extended ZsRE\cite{Yao, knowedit} and $\text{Wiki}_{counterfact}$ \cite{ripple-effects, knowedit}. Generally, they adopt four metrics: reliability, generality, portability, and locality. Given an initial base model $f_{\theta}$, a post-edit model $f_{\theta'}$, and a set of edit instances $(x_t,y_t)\in \{(x_t,y_t)\}$, the reliability is computed as the average accuracy of the edit cases:
\begin{equation}
\mathbb{E}_{(x_t,y_t)\in \{(x_t,y_t)\}} \{ \arg\max\nolimits_y f_{\theta'}(y|x_t) = y_t \} \ .
\end{equation}
The editing should also edit the equivalent neighbor of the instance $(x_t',y_t') \in N(x_t,y_t)$ (\textit{e.g.} rephrased descriptions). This metric is named generality and is evaluated by the average accuracy on the neighbors of the edit cases:
\begin{equation}
\mathbb{E}_{(x_t',y_t')\in \{N(x_t,y_t)\}} \{ \arg\max\nolimits_y f_{\theta'}(y|x_t') = y_t' \} \ .
\end{equation}
Beyond simple rephrasing, the editing is also supposed to affect other sophisticatedly related instances $(x_t'',y_t'') \in P(x_t,y_t)$. For example, instances that require reasoning, logical generalization over the edits. This metric is defined as portability:
\begin{equation}
\mathbb{E}_{(x_t'',y_t'')\in \{P(x_t,y_t)\}} \{ \arg\max\nolimits_y f_{\theta'}(y|x_t'') = y_t'' \} \ .
\end{equation}
Despite the editing, those instances that are irrelevant to the edit cases $(\hat{x_t},\hat{y_t}) \in \{O(x_t,y_t), f_{\theta}(x_t) = y_t\}$ should not be affected. This evaluation is called locality (also known as specificity) and is measured by the proportion of unchanged predictions between the initial model and the post-edit model:
\begin{equation}
\mathbb{E}_{(\hat{x_t},\hat{y_t})\in \{O(x_t,y_t)\}} \{ f_{\theta'}(\hat{x_t}) = f_{\theta}(\hat{x_t}) \} \ .
\end{equation}
For the extended ZsRE \cite{Yao, knowedit} and $\text{Wiki}_{counterfact}$ \cite{ripple-effects, knowedit}, we follow the setting in the original paper and combine reliability and generality to the Edit Success rate.

\subsection{Baseline implementation details}
\label{appendix.baseline-details}
Unless otherwise specified, the baselines are implemented by using the EasyEdit framework \cite{EasyEdit}.

\paragraph{Fine-tuning}
We follow the procedure implemented in previous work \cite{rome, memit, Yao, knowedit} to fine-tune a specific layer from the model. We select layer 13 for Llama-3.2-1B and layer 27 for Qwen2.5-7B. For both models, we adopt the learning rate of $5e^{-4}$ and the number of optimization steps 25.

\paragraph{LoRA}
For both models, we use LoRA \cite{LoRA} to update the query and key projection matrix of the models, with rank set to 8, $\alpha$ set to 32, the dropout rate 0.1, and the learning rate $5e^{-3}$. The number of updating steps is set to 70 for Llama-3.2-1B and 60 for Qwen2.5-7B.

\paragraph{ROME}
ROME \cite{rome} treats the FFN part of the LLMs as a key-value association and updates a pre-located layer by directly inserting an optimized key-value pair. We update the layer 5 for both Llama-3.2-1B and Qwen2.5-7B, and adopt 25 optimization steps for Llama-3.2-1B and 20 optimization steps for Qwen2.5-7B, with both learning rate $5e^{-5}$.

\paragraph{R-ROME}
R-ROME \cite{r-rome} is another version of ROME \cite{rome} with modified code implementation. We use the same hyperparameters as ROME.

\paragraph{KN}
KN \cite{kn} hypothesize that factual knowledge is stored in FFN memories and expressed by knowledge neurons. For both models, we use the threshold of 0.2 for knowledge attribution scores and 0.4 for the threshold of the prompts sharing percentage.

\paragraph{GRACE}
GRACE \cite{grace} adopts a discrete codebook to memorize the edits as key-value pairs. We set the location of the codebook layer 13 and 18 for Llama-3.2-1B and Qwen2.5-7B, respectively. Surprisingly, the $\epsilon$ value used in the original paper (1-3) seems insufficient for the complex editing experiments in this paper. Therefore, we increase it to 50. The number of optimization steps for the value vector is set to 100.

\paragraph{IKE}
IKE \cite{ike} maintains an explicit memory for edits and retrieves them via K-nearest neighbors. The retrieved edits are then used to construct demonstrations, which are then prefixed to the input to edit the behavior. In the experiments, we set $K=16$.

\paragraph{MEND}
MEND \cite{mend} trains an additional meta-network to predict a new rank-one update to the input gradient. In this paper, we train each model using ZsRE \cite{zsre} and COUNTERFACT \cite{memit}, and adopt the ZsRE-trained model to MQuAKE and the extended ZsRE and COUNTERFACT-trained model for $\text{Wiki}_{counterfact}$.

\paragraph{SERAC}
SERAC \cite{serac} employs an explicit edit-instance memory, an additional trained scope classifier, and a trained counterfactual model. The scope classifier is responsible for determining whether an input is relevant to the edits in the memory. The input is fed to the counterfactual model once the input is deemed as relevant to memorized edits and the original model otherwise. We use the distilbert-base-cased \cite{distil-bert} model as the scope classifier, and train it using the training set of ZsRE \cite{zsre} and COUNTERFACT \cite{memit} from EasyEdit \cite{EasyEdit}. Following \cite{dune}, we use instruction-tuned models (Llama-3.2-1B-Instruct\footnote{https://huggingface.co/meta-llama/Llama-3.2-1B-Instruct} and Qwen2.5-0.5B\footnote{https://huggingface.co/Qwen/Qwen2.5-0.5B-Instruct}) for the counterfactual model.

\paragraph{MEMIT}
MEMIT \cite{memit} is the extension of ROME \cite{rome} that supports a batch of edits at a time. Unlike ROME, which only updates a single pre-located layer, MEMIT spreads the update to a set of identified layers. We apply changes to layers 4-8 for both models. Following the settings in the original paper, we set $\lambda$, the hyperparameter that balances the weighting of new and old associations, to $1.5\times10^4$.

\paragraph{EMMET}
EMMET \cite{emmet} is an unification of ROME \cite{rome} and MEMIT \cite{memit}. Similar to ROME, we edit the layer 5 for both models, and set $\lambda=1e^5$.

\paragraph{RAG}
We adopt bge-base-en-v1.5 \cite{bge-base-en-v1.57} as our retriever. For batch editing, we treat the corresponding batch number of edits as our corpus, and retrieve the 10 most relevant edits for each testing query.

\paragraph{InComeS}
We apply cross-attention operations only for the second half of the model's layers since we found that the gist KV cache from the first half is not informative enough to allow effective edit selection. For the calculation of cross-attention during inference, we adopt a temperature of $T=0.45$ to the logits before softmax, which is found to be helpful for effective editing.

\section{Further Analysis}
\label{appendix.further-analysis}

\subsection{RAG vs. InComeS}
\label{appendix.RAG-InComeS}

\paragraph{Problem settings}
InComeS and RAG target different problems. In RAG, the query is given and used to retrieve the relevant documents via a retrieval model before decoding. However, InComeS does not make such an assumption, and it "edits" the model with the provided knowledge before seeing any actual queries.

\paragraph{Methodology}
From the perspective of methodology, InComeS conducts selection mechanisms in a way that is better integrated into the LM and with a finer granularity compared to RAG methods. First, RAG inputs the full query for retrieval to select relevant documents, and this process needs an extra retrieval model; in contrast, our method requires no extra models or retrieval steps and directly integrates the selection mechanism into our base LM. Moreover, our method dynamically performs selection for each token individually, which has a much finer granularity than the query-based selection in RAG. This supports a wider range of applications.

\paragraph{Experiments on Multi-hop edits}
Despite the differences between our method and RAG, we compare our method with RAG to demonstrate our method's superiority over the complex editing scenarios (Table \ref{tab:mquake-rag}). InComeS outperforms RAG in almost all cases for both single editing and batch editing. Note that the result of RAG is the same as the result of ICL in single editing, which does not need retrieval. We also conduct a scaling experiment to verify the effectiveness of InComeS over RAG in long context or massive edits scenario. The results (Table \ref{tab:incomes-rag-scaling}) show that the InComeS consistently outperforms RAG in different context lengths (number of edits).

\begin{table}[t]
\footnotesize
    \centering
    \setlength{\tabcolsep}{1.5mm}{
    \begin{adjustbox}{max width=\textwidth}
    \begin{tabular}{lc}
    \toprule
     Method & MquAKE \\
     \midrule
     \multicolumn{2}{c}{Llama-3.2-1B} \\
     \midrule
     InComeS - $12k$ & \textbf{49.55} \\
     RAG - $12k$ & 41.34 \\
     InComeS - $60k$ & \textbf{45.63} \\
     RAG - $60k$ & 39.29 \\
     InComeS - $120k$ & \textbf{40.38} \\
     RAG - $120k$ & 37.98 \\
     \midrule
     \multicolumn{2}{c}{Qwen2.5-7B} \\
     \midrule
     InComeS - $12k$ & \textbf{49.37} \\
     RAG - $12k$ & 43.27 \\
     InComeS - $60k$ & \textbf{44.12} \\
     RAG - $60k$ & 41.41 \\
     InComeS - $120k$ & \textbf{42.29} \\
     RAG - $120k$ & 39.86 \\
     \midrule
    \bottomrule
    \end{tabular}
    \end{adjustbox}}
    \caption{Scaling experiments for InComeS and RAG.}
    \label{tab:incomes-rag-scaling}
\end{table}

\paragraph{Implicit retrieval vs. Explicit retrieval}
Some may claim that the explicitness of the retrieval process in RAG is an advantage over InComeS. We argue that it is not hard to transfer the implicit retrieval process of InComeS to an explicit one. As the setting of our method strictly follows one edit - one gist (Figure \ref{fig.method}), the mapping between the text and its gist representations would not be hard to build. 

\begin{table*}[t]
\footnotesize
    \centering
    \setlength{\tabcolsep}{1.5mm}{
    \begin{adjustbox}{max width=\textwidth}
    \begin{tabular}{lccccccccc}
    \toprule
    \multirow{2}{*}{Method} & \multirow{2}{*}{Model} & \multicolumn{4}{c}{Single Editing} & \multicolumn{4}{c}{Batch Editing} \\ 
    \cmidrule(lr){3-6}\cmidrule(lr){7-10}
    
     %\cmidrule(lr){2-4}\cmidrule(lr){5-7}\cmidrule(lr){8-10}\cmidrule(lr){11-13}
     & & 2-edits & 3-edits & 4-edits & Avg & 2-edits & 3-edits & 4-edits & Avg \\
     \midrule
     Base & \multirow{4}{*}{Llama-3.2-1B} & 41.79 & 43.51 & 31.58 & 38.96 & 41.79 & 43.51 & 31.58 & 38.96 \\
     RAG && 59.23 & 59.00 & 51.63 & 56.62 & 40.65 & 46.78 & 43.58 & 43.67 \\
     ICL && 59.23 & 59.00 & 51.63 & 56.62 & 50.06 & 49.86 & 42.37 & 47.43 \\
     \textbf{InComeS} && \textbf{71.19} & \textbf{72.17} & \textbf{72.62} & \textbf{71.99} & \textbf{53.93} & \textbf{52.79} & \textbf{52.73} & \textbf{53.15} \\
     \midrule
     Base & \multirow{4}{*}{Qwen2.5-7B} & 44.08 & 44.14 & 30.62 & 39.61 & 44.08 & 44.14 & 30.62 & 39.61 \\
     RAG && 69.76 & 76.91 & 74.54 & 73.74 & 42.91 & 46.37 & 46.15 & 45.14 \\
     ICL && \textbf{69.76} & \textbf{76.91} & 74.54 & \textbf{73.74} & 53.53 & 50.54 & 44.77 & 49.61 \\
     \textbf{InComeS} && 66.46 & 71.24 & \textbf{76.54} & 71.41 & \textbf{55.13} & \textbf{53.48} & \textbf{47.91} & \textbf{52.17} \\ 
     \midrule
     Base & \multirow{4}{*}{Qwen3-8B-base} & 43.49 & 41.01 & 26.32 & 36.94 & 43.49 & 41.01 & 26.32 & 36.94 \\
     RAG && 55.82 & 54.13 & 47.06 & 52.34 & 45.98 & 46.52 & 46.67 & 46.39 \\
     ICL && 55.82 & 54.13 & 47.06 & 52.34 & 50.78 & 48.84 & 45.28 & 48.30 \\
     \textbf{InComeS} && \textbf{56.36} & \textbf{58.71} & \textbf{59.93} & \textbf{58.33} & \textbf{53.42} & \textbf{50.90} & \textbf{50.36} & \textbf{51.56} \\
     \midrule
    \bottomrule
    \end{tabular}
    \end{adjustbox}}
    \caption{Results for RAG on MQuAKE \cite{mquake}.}
    \label{tab:mquake-rag}
\end{table*}

\subsection{Side effect}
\label{appendix.side-effect}
We present a side effect analysis of our method in this section (Table \ref{tab:side-effect}). We test the editing side effect under three different numbers of edits ($0k$, $0.1k$, $1k$) on MMLU \cite{mmlu} benchmark, which consists of 57 tasks across 4 domains, namely Social science, Humanities, STEM, and others. The results indicate that increasing the number of edits does not significantly harm the model's general capability (lines 3 - 5 and 7 - 9 in Table \ref{tab:side-effect}), demonstrating the potential scalability of our method. The continuous pre-training brings an inevitable modest side effect to the model (lines 2 - 3 and 6- 7 in Table \ref{tab:side-effect}).

\begin{table*}[t]
\footnotesize
    \centering
    \setlength{\tabcolsep}{1.5mm}{
    \begin{adjustbox}{max width=\textwidth}
    \begin{tabular}{lccccccc}
    \toprule
    Method & Model & Social Sciences & Humanities & STEM & Other & Average \\ 
     \midrule
     Base & \multirow{4}{*}{Llama-3.2-1B} & 25.09 & 27.31 & 25.87 & 27.49 & 26.44 \\
     InComeS - w/ no edits && 23.18 & 25.91 & 22.66 & 26.06 & 24.45 \\
     InComeS - w/ $0.1k$ edits && 22.45 & 25.01 & 22.66 & 24.62 & 23.68 \\
     InComeS - w/ $1k$ edits && 22.91 & 25.30 & 22.73 & 24.11 & 23.76 \\
     \midrule
     Base & \multirow{4}{*}{Qwen2.5-7B} & 72.72 & 69.19 & 63.42 & 69.72 & 68.76 \\
     InComeS - w/ no edits && 68.06 & 65.68 & 59.58 & 66.39 & 64.93 \\
     InComeS - w/ $0.1k$ edits && 64.64 & 60.65 & 56.84 & 63.39 & 61.38 \\
     InComeS - w/ $1k$ edits && 63.86 & 62.08 & 57.58 & 62.88 & 61.60 \\
     \midrule
    \bottomrule
    \end{tabular}
    \end{adjustbox}}
    \caption{Side effect evaluation on MMLU \cite{mmlu}.}
    \label{tab:side-effect}
\end{table*}

\subsection{Efficiency analysis}
\label{appendix.further-efficiency}
Compared to many traditional editing methods that require model backward calculation, our method only requires one single forward pass for each editing context. In comparison to ICL, which needs to encode the entire concatenated edit context, our approach enables parallel encoding of multiple edits, leading to great efficiency gains for the encoding (prefilling) stage. In addition, the compressed context also accelerates the decoding phase compared to the ICL decoding with prefilled KV cache.
Here, we provide an analysis of our method's efficiency advantage over traditional ICL for both the encoding and decoding stages.

\paragraph{Encoding}
Assume we have $N$ edits and each edit has a Length of $L$. For ICL prefilling, it has to encode the whole sequence with length $N \times L$. However, for InComeS, each edit is processed individually, and it encodes edits in parallel. In this case, it encodes a batch of $N$ edits with length $L$. Thanks for the highly optimized GPU parallel computation, such a feature approximately reduces the time consumption by $N$ times.

\paragraph{Decoding}
Suppose we have $N$ compressed gists, which corresponds to $N$ individual edits with length $L$ and $N \times L$ tokens whose KV caches have been prefilled. For each decoding position, the ICL self-attention needs to compute a matrix with size $1 \times N \times L$. However, InComeS only needs to calculate the gist cross-attention matrix with size $1 \times N$. This roughly accelerates the decoding by $L$ times.

\subsection{Edit success \& Locality}
\label{appendix.edit-success-metric}
As expected, Table \ref{tab:portability-result-appendix} shows traditional editing methods such as fine-tuning and ROME exhibit high edit success rates; however, their portability scores generally lag behind the top-performing methods, highlighting a common limitation of these approaches. In contrast, ICL-based approaches that leverage the in-context learning capabilities of LLMs demonstrate superior performance in complex editing scenarios that require reasoning, owing to the enhanced context understanding of LLMs. We argue that the edit success metric may not be a comprehensive metric to evaluate the real capability of the post-edited model, as it can be cheated by the overfitting problem \cite{edit-overfit}, i.e., the model can assign disproportionately high probabilities to the edit target to get a high edit success rate. This may explain why the simple fine-tuning method shows an extremely high edit success rate but fails to maintain it on the porability. Therefore, we focus on complex settings, including multi-hop editing, natural language editing, and portability that aligns more with the real application scenarios. Our method demonstrates excellent performance in these settings, which aligns with our motivations described in section \ref{introduction}.

Our method shows great locality gain compared to ICL, which aligns with the results in the side effect analysis (Appendix \ref{appendix.side-effect}). This results from the inclusion of the zero gist, which flexibly allows the token to ignore the gist information if none of them is deemed relevant.

\subsection{Effectiveness of the method over model scale}
\label{appendix.model-scale-results}
We observe that the performance gain in Table \ref{tab:mquake} appears to diminish as the model scale increases. In this section, we conduct further experiments to find out whether this is true. The results in Table \ref{tab:model-scale-verification} show that the performance gain does not decrease when the model scale increases.

\subsection{Further scaling up contexts}
\label{appendix.further-scaling-up-contexts}
Although the performance gap between ICL and InComeS is already noticeable under $1k$ edits (section \ref{scale-up-contexts}), we conduct further experiments on Llama-3.2-1B to verify the effectiveness of InComeS under $10k$ edits, which is around $120k$ in context length. The results are shown in Table \ref{tab:scales-up-contexts-more}. InComeS shows consistent superiority over ICL.

\begin{table}[t]
\footnotesize
    \centering
    \setlength{\tabcolsep}{1.5mm}{
    \begin{adjustbox}{max width=\textwidth}
    \begin{tabular}{lcc}
    \toprule
    \multirow{2}{*}{Method} & \multicolumn{2}{c}{COUNTERFACT} \\ 
    \cmidrule(lr){2-3}
    & Reliability & Generality \\ 
     \midrule
     \multicolumn{3}{c}{$1k$ (Length around $12k$)} \\
     \midrule
     - ICL & 42.38 & 11.83 \\
     - InComeS & \textbf{75.82} & \textbf{23.44} \\
     \midrule
     \multicolumn{3}{c}{$5k$ (Length around $60k$)} \\
     \midrule
     - ICL & 37.85 & 10.88 \\
     - InComeS & \textbf{69.96} & \textbf{20.98} \\
     \midrule
     \multicolumn{3}{c}{$10k$ (Length around $120k$)} \\
     \midrule
     - ICL & 35.55 & 09.76 \\
     - InComeS & \textbf{60.76} & \textbf{16.77} \\
     \midrule
    \bottomrule
    \end{tabular}
    \end{adjustbox}}
    \caption{Further scales up contexts.}
    \label{tab:scales-up-contexts-more}
\end{table}

\subsection{Memory bound quantification}
\label{appendix.memory-bound-quantification}
During the inference stage, the additional memory introduced by InComeS is the storage of the KV cache of the gist representations. Take the Llama-3.2-1B as an example, if we have $1k$ edits, the memory required (in float16) for a single layer is $2 \times 2048 \times 1000 = 3.91MB$. Because the inference process is conducted sequentially, we can load one layer KV cache at a time. Therefore, 3.91MB is the extra memory required for inference. For the whole storage, since InComeS only use deeper half layers, the memory is $3.91 \times 8 = 31.28MB$. If the edit number goes considerably high, techniques like top-k can be easily applied to our method.

\begin{table}[t]
\footnotesize
    \centering
    \setlength{\tabcolsep}{1mm}{
    \begin{adjustbox}{max width=\linewidth}
    \begin{tabular}{lcc}
    \toprule
    Method & Single Edit & Batch Edit \\ 
     \midrule
     \multicolumn{3}{c}{Llama-3.2-1B} \\
     \midrule
     % Base & 38.96 & 38.96\\
     FT-M & 45.22 & 38.96 \\
     LoRA \cite{LoRA} & 54.23 & 40.84 \\
     ROME \cite{rome} & 3.43 & - \\
     % R-ROME \cite{r-rome} & 5.18 & - \\
     MEMIT \cite{memit} & 28.23 & 20.84 \\
     % EMMET \cite{emmet} & 5.58 & 14.85 \\
     % GRACE \cite{grace} & 6.94 & 2.26 \\
     % SERAC \cite{serac} & 38.97 & 39.01 \\
     MEND \cite{mend} & 27.95 & 23.99 \\
     RECIPE \cite{recipe} & 48.45 & 39.74 \\
     % MELO \cite{melo} & 40.33 & 41.47 \\
     ATBIAS \cite{atbias}  & 51.33 & 38.73 \\
     DR-IKE \cite{dr-ike} & 35.49 & 31.73 \\
     AlphaEdit \cite{alphaedit} & 44.74 & 35.63 \\
     ICE \cite{ice} & 40.74 & 31.23 \\
     ICL & 48.22 & 42.38 \\
     \textbf{InComeS} & \textbf{62.74} & \textbf{47.98} \\
     \midrule
     \multicolumn{3}{c}{Qwen2.5-7B} \\
     \midrule
     % Base & 39.61 & 39.61 \\
     FT-M & \textbf{61.44} & 37.88 \\
     LoRA \cite{LoRA} & 28.22 & 18.77 \\
     ROME \cite{rome} & 5.44 & - \\
     % R-ROME \cite{r-rome} & 7.01 & - \\
     MEMIT \cite{memit} & 30.99 & 27.12 \\
     % EMMET \cite{emmet} & 30.28 & 39.91 \\
     % GRACE \cite{grace} & 14.15 & 5.85 \\
     % SERAC \cite{serac} & 56.15 & 40.69 \\
     MEND \cite{mend} & 29.86 & 24.77 \\
     RECIPE \cite{recipe} & 50.74 & 39.51 \\
     % MELO \cite{melo} & 40.49 & 42.20 \\
     ATBIAS \cite{atbias} & 41.11 & 37.00 \\
     DR-IKE \cite{dr-ike} & 44.14 & 33.13 \\
     AlphaEdit \cite{alphaedit} & 48.21 & 39.49 \\
     ICE \cite{ice} & 55.97 & 38.11 \\
     ICL & 60.78 & 37.63 \\
     \textbf{InComeS} & 61.42 & \textbf{42.29} \\
     \midrule
    \bottomrule
    \end{tabular}
    \end{adjustbox}}
    \caption{Results on MQuAKE \cite{mquake} for wild evaluation framework \cite{wild}.}
    \label{tab:mquake-wild}
\end{table}

\subsection{Meta training cost analysis}
The meta-training cost is smaller than a simple SFT these days. Take Olmo-3-7B-Instruct-SFT as an example; the data\footnote{https://huggingface.co/datasets/allenai/Dolci-Instruct-SFT} used for fine-tuning this model has at least 4B tokens, which is far larger than our meta-training cost (1.5B tokens). Given that the fine-tuning is a very common stage employed by almost all models nowadays, we believe such cost is not a considerable barrier for practical deployment.

Furthermore, compared to some tranditional light-weight editing method like MEMIT \cite{memit}:
\begin{itemize}
    \item InComeS has better performance in almost all scenarios reported in the paper.
    \item InComeS has better flexibility and supports the edits in any \textbf{natural language form}, while MEMIT-like optimization requires the edits follow a \textbf{triplet-like structure}, which is unrealistic for deployment.
    \item InComeS requires meta-training for different base models, but MEMIT \cite{memit} also needs to compute the covariance matrix before editing for different models.
\end{itemize}

Compared to ICL:
\begin{itemize}
    \item InComeS has better performance in almost all scenarios reported in the paper.
    \item InComeS requires an additional meta-training, but acquires a significant speedup for future inference (Table \ref{tab:time-analysis-encoding-decoding}). We believe this is a great gain for future practical deployment.
\end{itemize}

\begin{table}[t]
\footnotesize
    \centering
    \setlength{\tabcolsep}{1mm}{
    \begin{adjustbox}{max width=\linewidth}
    \begin{tabular}{lcc}
    \toprule
    Method & ZsRE-extended & $\text{Wikidata}_{counterfact}$ \\ 
     \midrule
     \multicolumn{3}{c}{Llama-3.2-1B} \\
     \midrule
     % Base & 38.96 & 38.96 \\
     FT-M & 80.99 / 78.41 & 48.71 \\
     LoRA \cite{LoRA} & 81.22 / 77.64 & 78.23 / 71.39 \\
     ROME \cite{rome} & 72.97 / - & 75.88 / - \\
     % R-ROME \cite{r-rome} & 5.18 & - \\
     MEMIT \cite{memit} & 70.33 / 53.41 & 67.43 / 53.27 \\
     % EMMET \cite{emmet} & 5.58 & 14.85 \\
     % GRACE \cite{grace} & 6.94 & 2.26 \\
     % SERAC \cite{serac} & 38.97 & 39.01 \\
     MEND \cite{mend} & 31.26 / 29.34 & 27.66 / 21.78 \\
     % MELO \cite{melo} & 40.33 & 41.47 \\
     ICL & 66.74 / 55.73 & 84.39 / 68.88 \\
     \textbf{InComeS} & \textbf{85.21 / 78.88} & \textbf{85.31 / 70.84} \\
     \midrule
     \multicolumn{3}{c}{Qwen2.5-7B} \\
     \midrule
     % Base & 39.61 & 39.61 \\
     FT-M & 78.99 / 74.41 & 88.23 / 79.41 \\
     LoRA \cite{LoRA} & 76.45 / 62.37 & 66.41 / 57.29 \\
     ROME \cite{rome} & 78.37 / - & 79.41 / - \\
     % R-ROME \cite{r-rome} & 7.01 & - \\
     MEMIT \cite{memit} & 74.31 / 63.21 & 80.36 / 78.31 \\
     % EMMET \cite{emmet} & 30.28 & 39.91 \\
     % GRACE \cite{grace} & 14.15 & 5.85 \\
     % SERAC \cite{serac} & 56.15 & 40.69 \\
     MEND \cite{mend} & 35.18 / 25.32 & 39.75 / 26.77 \\
     % MELO \cite{melo} & 40.49 & 42.20 \\
     ICL & 67.88 / 58.76 & 81.45 / 70.57 \\
     \textbf{InComeS} & 88.34 / 80.41 & \textbf{82.36 / 71.43} \\
     \midrule
    \bottomrule
    \end{tabular}
    \end{adjustbox}}
    \caption{Results on ZsRE-extended and $\text{Wikidata}_{counterfact}$ for wild evaluation framework \cite{wild}.}
    \label{tab:zsre-wikidata-wild}
\end{table}

\begin{table*}[t]
\footnotesize
    \centering
    \setlength{\tabcolsep}{1.5mm}{
    \begin{adjustbox}{max width=\textwidth}
    \begin{tabular}{lccccccccc}
    \toprule
    \multirow{2}{*}{Layers} & \multirow{2}{*}{Model} & \multicolumn{4}{c}{Single Editing} & \multicolumn{4}{c}{Batch Editing} \\ 
    \cmidrule(lr){3-6}\cmidrule(lr){7-10}
    
     & & 2-edits & 3-edits & 4-edits & Avg & 2-edits & 3-edits & 4-edits & Avg \\
     \midrule
     Base & \multirow{6}{*}{Qwen2.5-7B} & 44.08 & 44.14 & 30.62 & 39.61 & 44.08 & 44.14 & 30.62 & 39.61 \\
     28 && 45.57 & 46.48 & 32.23 & 41.43 & 43.12 & 42.31 & 32.87 & 39.43 \\ 
     24-28 && 55.12 & 62.74 & 61.13 & 59.66 & 49.45 & 44.32 & 39.86 & 44.54 \\
     20-28 && 55.96 & 61.95 & 63.23 & 60.38 & 50.35 & 45.63 & 40.33 & 45.46 \\
     17-28 && 62.87 & 68.95 & 69.64 & 67.15 & 53.26 & 51.38 & 45.77 & 50.14 \\
     14-28 (InComeS) && \textbf{66.46} & \textbf{71.24} & \textbf{76.54} & \textbf{71.41} & \textbf{55.13} & \textbf{53.48} & \textbf{47.91} & \textbf{52.17} \\
     \midrule
    \bottomrule
    \end{tabular}
    \end{adjustbox}}
    \caption{Analysis on the effectiveness of the deeper half layers.}
    \label{tab:analysis-on-deeper-half-layers}
\end{table*}

\begin{table*}[t]
\footnotesize
    \centering
    \setlength{\tabcolsep}{1.5mm}{
    \begin{adjustbox}{max width=\textwidth}
    \begin{tabular}{lccccccccc}
    \toprule
    \multirow{2}{*}{Method} & \multirow{2}{*}{Model} & \multicolumn{4}{c}{Single Editing} & \multicolumn{4}{c}{Batch Editing} \\ 
    \cmidrule(lr){3-6}\cmidrule(lr){7-10}
    
     %\cmidrule(lr){2-4}\cmidrule(lr){5-7}\cmidrule(lr){8-10}\cmidrule(lr){11-13}
     & & 2-edits & 3-edits & 4-edits & Avg & 2-edits & 3-edits & 4-edits & Avg \\
     \midrule
     Base & \multirow{3}{*}{Llama-3.2-1B} & 41.79 & 43.51 & 31.58 & 38.96 & 41.79 & 43.51 & 31.58 & 38.96 \\
     ICL && 59.23 & 59.00 & 51.63 & 56.62 & 50.06 & 49.86 & 42.37 & 47.73 \\
     \textbf{InComeS} && \textbf{71.19} & \textbf{72.17} & \textbf{72.62} & \textbf{71.99} & \textbf{53.93} & \textbf{52.79} & \textbf{52.73} & \textbf{53.15} \\
     \midrule
     Base & \multirow{3}{*}{Llama-3.1-8B} & 45.29 & 45.69 & 31.66 & 40.88 & 45.29 & 45.69 & 31.66 & 40.88 \\
     ICL && 58.76 & 56.99 & 45.47 & 53.74 & 48.64 & 49.58 & 39.51 & 45.91 \\
     \textbf{InComeS} && \textbf{74.39} & \textbf{78.25} & \textbf{75.24} & \textbf{75.96} & \textbf{56.28} & \textbf{56.55} & \textbf{51.90} & \textbf{54.91} \\
     \midrule
    \bottomrule
    \end{tabular}
    \end{adjustbox}}
    \caption{Results for different model scale \cite{mquake}.}
    \label{tab:model-scale-verification}
\end{table*}

\subsection{More results}
\label{appendix.more-results-on-qwen3}
To further verify the effectiveness of our method over the recent model, we further test our method on Qwen3-8B-base. The results (Table \ref{tab:dune-results-more} and Table \ref{tab:mquake-results-more}) demonstrate that our method works well in recent models. We also perform an evaluation using the WILD framework \cite{wild}, the results are shown in Tables \ref{tab:mquake-wild} and \ref{tab:zsre-wikidata-wild}.

\subsection{More related work}
\label{more-related-work}
Recent advances focus on improving the efficiency and expressiveness of soft prompting. ADePT \cite{adept} decomposes the soft prompt into a shorter prompt and a token-shared feed-forward network, enabling adaptive embedding offsets that vary according to model input, consistently outperforming prior methods across 23 NLP tasks. CE-Prompt \cite{ce-prompt} introduces composite embeddings via multiple randomly initialized embedding layers, enhancing prompt expression stability and significantly reducing inference computational costs. \citet{MuppidiNB25} propose Input-Dependent Soft Prompting with a self-Attention Mechanism (ID-SPAM), which generates soft prompts dynamically based on input tokens, demonstrating improved zero-shot domain transfer capability. Position also plays a critical role: \citet{Position-Really-Matters} provides a theoretical analysis showing that optimizing the prompt's position to surround the input can capture additional semantic information beyond traditional prefix or postfix methods.

\begin{table*}[t]
\footnotesize
    \centering
    \setlength{\tabcolsep}{1.5mm}{
    \begin{adjustbox}{max width=\textwidth}
    \begin{tabular}{lccccccccc}
    \toprule
    \multirow{2}{*}{Method} & \multirow{2}{*}{Model} & \multicolumn{4}{c}{Single Editing} & \multicolumn{4}{c}{Batch Editing} \\ 
    \cmidrule(lr){3-6}\cmidrule(lr){7-10}
    
     %\cmidrule(lr){2-4}\cmidrule(lr){5-7}\cmidrule(lr){8-10}\cmidrule(lr){11-13}
     & & 2-edits & 3-edits & 4-edits & Avg & 2-edits & 3-edits & 4-edits & Avg \\
     \midrule
     Base & \multirow{6}{*}{Qwen3-8B-base} & 43.49 & 41.01 & 26.32 & 36.94 & 43.49 & 41.01 & 26.32 & 36.94 \\
     FT-M && 51.32 & 50.39 & 45.22 & 48.98 & 48.11 & 44.24 & 41.39 & 44.58 \\
     LoRA \cite{LoRA} && 38.95 & 36.27 & 33.02 & 36.08 & 24.41 & 21.24 & 22.12 & 22.59 \\
     MEMIT \cite{memit} && 49.14 & 46.05 & 41.15 & 45.45 & 42.34 & 40.10 & 33.45 & 38.63 \\
     ICL && 55.82 & 54.13 & 47.06 & 52.34 & 50.78 & 48.84 & 45.28 & 48.30 \\
     \textbf{InComeS} && \textbf{56.36} & \textbf{58.71} & \textbf{59.93} & \textbf{58.33} & \textbf{53.42} & \textbf{50.90} & \textbf{50.36} & \textbf{51.56} \\
     \midrule
    \bottomrule
    \end{tabular}
    \end{adjustbox}}
    \caption{More results on mquake \cite{mquake}.}
    \label{tab:mquake-results-more}
\end{table*}

\begin{table*}[t]
\footnotesize
    \centering
    \setlength{\tabcolsep}{1.5mm}{
    \begin{adjustbox}{max width=\textwidth}
    \begin{tabular}{lccccccccc}
    \toprule
    \multirow{2}{*}{Method} & \multirow{2}{*}{Model} & \multicolumn{4}{c}{Single Editing} & \multicolumn{4}{c}{Batch Editing} \\ 
    \cmidrule(lr){3-6}\cmidrule(lr){7-10}
    
     %\cmidrule(lr){2-4}\cmidrule(lr){5-7}\cmidrule(lr){8-10}\cmidrule(lr){11-13}
     & & new info & scientific & de-biasing & Avg & new info & scientific & de-biasing & Avg \\
     \midrule
     Base & \multirow{6}{*}{Qwen3-8B-base} & 82.99 & 86.62 & 31.07 & 66.89 & 82.99 & 86.62 & 31.07 & 66.89 \\
     FT-M && 78.23 & 75.44 & 56.38 & 70.02 & 76.33 & 72.34 & 54.28 & 67.65 \\
     LoRA \cite{LoRA} && 71.23 & 69.99 & 43.23 & 61.48 & 73.13 & 64.59 & 42.22 & 59.98 \\
     MEMIT \cite{memit} && 81.29 & 80.45 & 49.65 & 70.46 & 80.99 & 78.53 & 45.66 & 68.39 \\
     ICL && 83.06 & 81.12 & 45.45 & 69.88 & 83.35 & 82.76 & 30.34 & 65.48 \\
     \textbf{InComeS} && \textbf{83.17} & \textbf{85.36} & \textbf{63.96} & \textbf{77.41} & \textbf{84.17} & \textbf{87.57} & \textbf{65.96} & \textbf{79.23} \\
     \midrule
    \bottomrule
    \end{tabular}
    \end{adjustbox}}
    \caption{More results on dune \cite{dune}.}
    \label{tab:dune-results-more}
\end{table*}

\end{document}